\documentclass[10pt, a4paper, oneside]{article}

\usepackage[authoryear,square,sort]{natbib}
\setcitestyle{aysep={}}                    
\makeatletter
\renewcommand\@biblabel[1]{[#1]}
\let\jucs@oldbibitem\bibitem
\renewcommand{\bibitem}[2][]{%
  \jucs@oldbibitem[#1]{#2}%
  \ifx\relax#1\relax\else\jucs@inlinelabel{#1}\fi
}
\newcommand{\jucs@inlinelabel}[1]{%
  \def\jucs@tmp{#1}%
  \ifx\jucs@tmp\@empty\else
    \jucs@splitlabel#1(\@nil\@nil%
  \fi
}
\def\jucs@splitlabel#1(#2\@nil#3\@nil{%
  \def\jucs@auth{#1}%
  \jucs@extractyear#2\@nil%
  \leavevmode\textbf{[\jucs@auth\ \jucs@yr]}\hspace{0.6em}\ignorespaces
}
\def\jucs@extractyear#1#2#3#4#5\@nil{%
  \def\jucs@yr{#1#2#3#4}%
}
\makeatother
\usepackage[colorlinks=true, linkcolor=blue, citecolor=blue, urlcolor=blue]{hyperref}
\usepackage{jucs2e}          
\usepackage{graphicx}
\usepackage{url}
\usepackage{ulem}            
\usepackage{mathtools}
\usepackage{amsmath,amssymb}
\usepackage{scalerel}
\usepackage{setspace}
\usepackage[strict]{changepage}
\usepackage{caption}
\usepackage{afterpage}
\usepackage{ragged2e}
\usepackage{booktabs}
\usepackage{multirow}
\usepackage{array}
\usepackage{tabularx}
\usepackage{longtable}
\usepackage{float}
\usepackage{makecell}
\usepackage{algorithm}
\usepackage{algpseudocode}
\usepackage{etoolbox}  
\usepackage{multicol}
\usepackage{placeins}
\usepackage[activate={true,nocompatibility},final,tracking=true,kerning=true,spacing=true,factor=1100,stretch=15,shrink=15]{microtype}
\microtypecontext{spacing=nonfrench}

\usepackage[textwidth=12.2cm, left=4.6cm, right=4.2cm,
            top=3.7cm, bottom=6.6cm, a4paper,
            headheight=0.5cm, headsep=0.5cm]{geometry}

\usepackage{fancyhdr}
\usepackage[format=plain, labelfont=it, textfont=it,
            justification=centering]{caption}

\usepackage{mathptmx}

\usepackage{titlesec}
\titleformat{\section}{\large\bfseries}{\thesection}{1em}{}
\titleformat{\subsection}{\normalsize\bfseries}{\thesubsection}{1em}{}
\titleformat{\subsubsection}{\normalsize\bfseries}{\thesubsubsection}{1em}{}

\renewcommand{\arraystretch}{1.05}

\algrenewcommand\algorithmicrequire{\textbf{Input:}}
\algrenewcommand\algorithmicensure{\textbf{Output:}}
\AtBeginEnvironment{algorithmic}{%
  \setlength{\itemsep}{3pt}%
  \setlength{\parsep}{0pt}%
  \setlength{\topsep}{4pt}%
  \setlength{\partopsep}{0pt}%
  \linespread{1.0}\selectfont%
}

\newcommand\paperauthor{{Moattari, M.: }}
\newcommand\papertitle{Interpretable Multimodal Classification with Linear Discriminant Tree Ensembles}
\newcommand\startingPage{1}
\header{\paperauthor \papertitle}

\newcommand\jucs{{Journal of Universal Computer Science}}
\newcommand\jucsvol{vol. xx, no. x (2025)}
\newcommand\jucspages{xxxx-xxxx}
\newcommand\jucssubmitted{submitted: xx/xx/2025}
\newcommand\jucsaccepted{accepted: xx/xx/2025}
\newcommand\jucsappeared{appeared: xx/xx/2025}
\newcommand\jucslicence{CC BY 4.0}

\begin{document}

\title{{\fontsize{14pt}{14pt}\selectfont{%
\vspace*{-3mm}%
Interpretable Multimodal Classification with Linear Discriminant Tree Ensembles%
\vspace*{-1mm}}}}

\author{%
  {\bfseries\fontsize{10pt}{10pt}\selectfont{Mojtaba Moattari}} \\
  {\fontsize{9pt}{12pt}\selectfont{%
    (Independent Researcher, Shiraz, Iran \\
    ORCID: 0000-0001-6191-6467, \\
    moatary.m@gmail.com)}}
}

\maketitle
\label{first}

{\fontfamily{ptm}\selectfont
\begin{abstract}
{\fontsize{9pt}{9pt}\selectfont{\vspace*{-2mm}
Multimodal affect and behaviour classifiers that fuse heterogeneous text, audio, and visual streams must simultaneously achieve competitive accuracy and produce human-under\-stand\-able explanations of the cues driving their decisions---a dual objective that current high-capacity models, notably Transformers, only partially address. While Transformers attain strong predictive performance, their distributed representations and deep nonlinearity make it difficult to assign meaningful importance weights to individual multimodal features, limiting their use in trust-sensitive applications such as clinical affect monitoring and educational assessment. We address this gap by developing a framework based on tree-based ensembles that balances accuracy and interpretability. The framework encodes each modality into tokens, extracts and clusters concepts to reduce dimensionality, routes the fused modalities through tree-based ensemble classifiers, and interprets trends using a novel modified feature importance metric. The modified importance reduces the influence of the negative  class in binary classification tasks, thereby improving indicator or marker detection. We evaluate on IEMOCAP, CMU-MOSI, and a custom Multimodal Mathematics dataset, reporting accuracy, standard F1, and a modified F1-measure (F1-mod) that emphasizes positive-class performance. The proposed tree-based ensembles---Linear Discriminant Tree (LDT), Linear Discriminant Forest (LDF), and Linear Discriminant AdaBoost (LDAB)---achieve F1-mod gains of 4.3\% over the Multimodal Transformer~\citep{tsai2019multimodal} and accuracy gains of 3.0\% over the primary interpretable multimodal baseline, Interpretable Multimodal Routing (IMR)~\citep{tsai2020multimodal}. The proposed multimodal feature importance extracts salient inter-modal concepts with substantially higher human-annotator agreement scores than default feature importance (62.2\% vs.\ 43.2\% on IEMOCAP; 46.7\% vs.\ 32.1\% on CMU-MOSI). The framework also characterizes cross-modal concept associations, a capability that standard attention-based deep learning models do not readily provide.
}}
\end{abstract}}

{\fontfamily{ptm}\selectfont
\begin{keywords}
{\fontsize{9pt}{9pt}\selectfont{%
Oblique Trees; Ensemble Learning; Context Encoding; Feature Importance; Multimodal Interpretability; Multimodal Fusion}}
\end{keywords}}

{\fontfamily{ptm}\selectfont
\begin{category}
{\fontsize{9pt}{9pt}\selectfont{%
I.2.6, I.5.2, I.5.4, H.2.8}}
\end{category}}

{\fontfamily{ptm}\selectfont
\begin{doi}
{\fontsize{9pt}{9pt}\selectfont{%
10.3897/jucs.<SubmissionNumber>}}
\end{doi}}

\section{Introduction}
\label{sec:intro}

Recent advances in interpretable multimodal learning have enabled text--audio--video fusion systems capable of modelling affective, behavioural, and clinical indicators while generating human-under\-stand\-able explanations of the multimodal cues underlying diagnostic or behavioural inferences~\citep{picard2000behavior,poria2017review,pelachaud2009modeling,clore2001behavior}. Such explanations support both model fault diagnosis and actionable behavioural guidance for clinicians and end users.

Transformer-based models constitute the current state of the art in multimodal affect and behaviour classification~\citep{tsai2019multimodal,vaswani2017attention,zadeh2019factorized,sahay2020lowrank}, yet they are computationally demanding and offer limited transparency for classification, interpretation, and feature extraction~\citep{sun2017review,kovalerchuk2021survey,gao2020survey}. Decision Trees~\citep{song2015decision}, by contrast, have lower accuracy on large datasets but are inherently interpretable~\citep{molnar2020interpretable,kovalerchuk2021survey}; they therefore represent a promising foundation for building (ensemble) classifiers that are both competitive and explainable. Transformer-based models face two specific interpretability obstacles:
\begin{itemize}
  \item They are not interpretable by design; consequently, the features they identify as important may not faithfully reflect local instance-level behaviour~\citep{molnar2020interpretable,ribeiro2016model}.
  \item Their depth, distributed representations, and cross-modal attention mechanisms make direct interpretation substantially more difficult than with inherently transparent models such as decision trees.~\citep{devlin2018bert,nagrani2021attentionbottlenecks}.
\end{itemize}

Furthermore, recent attention-based classifiers are computationally intensive and do not represent whole-dataset trends in a readily interpretable form~\citep{ribeiro2016model,liang2021attention,gkoumas2021difference}. This limitation arises either from distributed connectionist representations, which prevent localised information encoding, or from complex multi-stage pipelines, which obscure which layer's weights should constitute cross-modal or global importance signals. Representative examples include MMPNet~\citep{song2025multimodal} (concept-prototype centred interpretability), 3WD-DRT~\citep{jiang20253wd} (reliability-aware multimodal gating), AtCAF~\citep{gandhi2024atcaf} (causal contextual explanation), and MISA~\citep{hazarika2020misa} (shared modality-specific multimodal explanation). Existing interpretable-by-design models---such as Oblique Decision Trees~\citep{murthy1994system}, Sparse Oblique Trees~\citep{hada2023sparse}, and Linear Discriminant Trees (LDTs)~\citep{yildiz2005linear}---are, respectively, context-insensitive, computationally expensive, and prone to overfitting the training data.

To overcome these limitations, we propose a novel pipeline and feature encoder for multimodal classification. We focus on ensembles of Decision Trees (DTs)~\citep{song2015decision,pal2005random,ying2013advances,an2010new} and augment them with inductive biases, feature encoders, and a modified interpretability metric. If tree-based ensemble classifiers can be made sufficiently expressive for multimodal settings, the resulting framework may offer both competitive accuracy relative to Transformer-based models~\citep{tsai2019multimodal,vaswani2017attention,zadeh2017tfn,zadeh2018mfn} and superior interpretability relative to interpretable neural alternatives~\citep{molnar2020interpretable,ribeiro2016model,kovalerchuk2021survey,kim2018interpretability}. To the best of our knowledge, no prior work systematically compares different Decision Tree ensemble types~\citep{murthy1994system,hada2023sparse,yildiz2005linear,lemmond2010extended} in a unified multimodal framework.

\paragraph{Contributions}
The principal contributions of this work are as follows, each addressing a gap identified above:
\begin{itemize}
  \item \textbf{LDA-node projection inside DT, RF, AB for multimodal fusion.} Inserting a Linear Discriminant Analysis (LDA) projector at every tree node is an unexplored configuration in ensemble learning. Prior oblique-tree work~\citep{murthy1994system,hada2023sparse} applies projection only within a single tree or uses PCA/ICA rather than a class-discriminative criterion. We extend this idea to multimodal DT, Random Forest (RF), and AdaBoost (AB), which collectively constitute the LDT, LDF, and LDAB variants.
  \item \textbf{Positive-class-focused feature importance.} Standard feature importance is dominated by nodes that reject negative instances, misrepresenting positive-class cues. Our modified importance up-weights nodes where positive instances dominate the branch, yielding more interpretable and fault-diagnosable explanations.
  \item \textbf{Three context-integration schemes for multimodal tokens.} We design and compare K-means clustering, eigenvalue-based binary hierarchical clustering, and multi-sense clustering as alternative strategies for encoding modality-specific tokens before fusion, and we demonstrate that the choice of scheme substantially affects both accuracy and interpretability.
  \item A tree-based ensemble framework that combines context encoding with an appropriate classifier variant to improve multimodal binary classification~\citep{gkoumas2021difference,snoek2005earlylate}.
  \item A three-modality visualisation approach across three context-embedding schemes for local (sample-level) interpretation~\citep{selvaraju2017grad,kim2018interpretability}.
\end{itemize}

Experimental results are obtained on IEMOCAP, CMU-MOSI, and the custom
Multimodal Mathematics dataset. The primary interpretable baseline throughout this
paper is IMR (Interpretable Multimodal Routing)~\citep{tsai2020multimodal};
comparisons with non-interpretable models such as MulT~\citep{tsai2019multimodal}
are secondary. The proposed ensembles achieve F1-mod gains of 4.3\% over MulT and
accuracy gains of 3.0\% over IMR. Ablation studies (Section~\ref{sec:ablation})
confirm that n-gram count correlates strongly with accuracy and F1 in non-sparse
configurations~\citep{chelba2017ngram}. Re\-pro\-duc\-ibil\-ity materials---including
code, preprocessed features, and trained models---are available at
\url{https://github.com/moatary/reproducible_multimodal_ensemble}. Furthermore,
all hyperparameter search spaces and final selected values are detailed in
Table~\ref{tab:hyperparams}.

\paragraph{Research questions} This work addresses the following research questions:
\begin{enumerate}
  \item \textbf{RQ1:} Can tree-based ensemble models~\citep{pal2005random,ying2013advances,an2010new,lemmond2010extended} achieve performance comparable to Transformer-based models~\citep{tsai2019multimodal,zadeh2019factorized,sahay2020lowrank} in multimodal settings?
  \item \textbf{RQ2:} Can modified feature importance improve multimodal interpretability---covering model fault diagnosis and condition detection---in both global and local settings~\citep{molnar2020interpretable,ribeiro2016model,kovalerchuk2021survey,kim2018interpretability}?
  \item \textbf{RQ3:} Does LDA-node projection in tree-based ensembles improve classification accuracy over standard tree-based ensembles~\citep{yildiz2005linear,lemmond2010extended}?
  \item \textbf{RQ4:} Does a specific context-clustering approach lead to measurably improved interpretability~\citep{kovalerchuk2021survey,kim2017bag,athiwaratkun2018probabilistic}?
\end{enumerate}

The remainder of this paper is organised as follows. Section~\ref{sec:related} reviews related work. Section~\ref{sec:methods} presents the proposed framework and feature importance derivation. Section~\ref{sec:results} reports results, comparisons, ablation studies, and interpretability analyses. Section~\ref{sec:conclusion} concludes and outlines future directions.

\section{Related Work}
\label{sec:related}

\subsection{Random Forest, Adaptive Booster, and Gradient Booster}

Random forests build an ensemble by training each tree on a bootstrap subsample of the data, with trees grown in parallel~\citep{pal2005random}. The final prediction is determined by majority vote. AdaBoost~\citep{ying2013advances,an2010new} iteratively reweights training instances, directing subsequent weak learners on previously misclassified examples. The final prediction aggregates weighted votes from all weak learners. XGBoost grows trees sequentially to minimise the ensemble's residual error, defining a complexity penalty as:
\begin{equation}
\Omega(f_t) = \gamma L_t + \tfrac{1}{2}\lambda \sum_{j=1}^{L_t} \omega_j^2
\label{eq:xgb_complexity}
\end{equation}
where $L_t$ is the number of leaves and $\omega_j$ is the score of leaf $j$.

\subsection{Linear Discriminant Tree (LDT)}

The Linear Discriminant Tree splits inputs using a linear projection extracted by Linear Discriminant Analysis (LDA), seeking oblique hyperplanes that minimise node impurity~\citep{yildiz2005linear,murthy1994system}. Fisher's LDA maximises the ratio of between-class to within-class variance. In the two-class case, the projection vector $\mathbf{w}$ and intercept $b$ are:
\begin{equation}
\mathbf{w} = \Sigma^{-1}(\boldsymbol{\mu}_1 - \boldsymbol{\mu}_0)
\end{equation}
\begin{equation}
b = -\tfrac{1}{2}(\boldsymbol{\mu}_1 + \boldsymbol{\mu}_0)^{\!\top}\Sigma^{-1}(\boldsymbol{\mu}_1 - \boldsymbol{\mu}_0)
\end{equation}
where $\Sigma^{-1}$ is the inverse pooled covariance matrix and $\boldsymbol{\mu}_k$ is the class-$k$ mean.

In this work, we extend LDA-based node projection to RF, AB, and XGBoost (XGB), and we build a novel feature importance on top of these projectors that weights each nodes by its positive-class contribution---a capability not available in standard or sparse oblique trees where a feature affects only a subset of nodes~\citep{hada2023sparse}.

\begin{figure}[t]
  \centering
  \scalebox{0.65}{\includegraphics[width=0.9\linewidth]{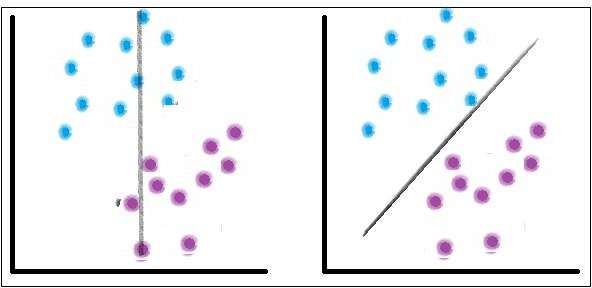}}
  \caption{\fontsize{10pt}{11pt}\selectfont{\itshape{Axis-aligned (left) vs.\ LDA-oblique (right) decision boundaries for a node split. The LDA hyperplane separates the two classes more cleanly with a lower misclassification count.}}}
  \label{fig:ldt-illustration}
\end{figure}

Algorithm~\ref{alg:node-optimizer} optimizes the data subset routed to each DT node by selecting the LDA-oblique boundary (Figure~\ref{fig:ldt-illustration}) that yields the highest impurity reduction.

\begin{algorithm}[t]
\caption{Node impurity optimiser}
\label{alg:node-optimizer}
\begin{algorithmic}[1]
\Require Discriminatory components/eigenvectors; dimensionality-reduced context data;
\Statex \hspace{2.2em}\texttt{dim}: number of node projector components; context labels
\Ensure Selected candidate component; node decision (positive or negative);
\Statex \hspace{2.2em}the $m$ features with highest and lowest component values
\For{each component}
  \State Find \texttt{dim} weight vectors that maximize the projector objective (Section~\ref{sec:node-projectors}).
  \State Select the composition with maximum impurity reduction.
  \State \texttt{argsort} the component projector.
  \State Record selected projector component and decision.
\EndFor
\end{algorithmic}
\end{algorithm}

\subsection{Contrast with Existing Oblique-Tree and Concept-Based Methods}

\paragraph{Oblique and LDA-based trees} Murthy et al.~\citet{murthy1994system} introduced oblique decision trees with multivariate splits but did not apply discriminant projectors inside ensemble learners. Hada et al.~\citet{hada2023sparse} proposed Sparse Oblique Trees that reduce computational cost through sparsity but remain single-tree models and do not leverage class-discriminative projections. Yildiz and Alpaydin~\citet{yildiz2005linear} proposed the LDT, yet restricted
it to a single stand-alone tree. Lemmond et al.~\citet{lemmond2010extended} extended discriminant random forests but did not generalise to boosting variants. Our framework is, to our knowledge, the first to embed LDA projectors inside DT, RF, and AB simultaneously, enabling systematic comparison under a unified multimodal pipeline.

\paragraph{Concept-based interpretability} Concept Activation Vectors (CAV)~\citep{kim2018interpretability} cluster convolutional receptive-field outputs but are not directly applicable to tokenized multimodal data. Concept-GradCAM~\citep{selvaraju2017grad} maps label-related activations to image regions yet is restricted to unimodal visual inputs. Bag of Concepts~\citep{kim2017bag} clusters word embeddings into a bag-of-words representation but provides no mechanism for multimodal concept fusion. Our framework addresses all three gaps: it constructs a bag of multimodal concepts, applies it across three modalities simultaneously, and derives importance weights from class-discriminative tree nodes rather than gradient-based attribution.

\subsection{Interpretability}

Standard tree feature importance heavily weights nodes that discriminate against the negative instances (false positives), which may distort importance scores for positive-class features and mislead practitioners~\citep{molnar2020interpretable}. Existing highly cited interpretability methods provide neither sample-level nor word-level global interpretation for multimodal data~\citep{ribeiro2016model,kim2018interpretability}. We design dimensionality-reduction-based modality visualisation and a contextual feature importance to address these gaps, using the linear node projector to weight each feature's contribution across all nodes~\citep{selvaraju2017grad,kim2018interpretability}.

\subsection{Accurate and Interpretable Models in Multimodal Fusion}

Hu et al.~\citet{huh2021interpretable} proposed multimodal fusion using canonical correlation and Grad-CAM~\citep{selvaraju2017grad} for neuroimaging. Liang et al.~\citet{liang2018multimodal} proposed a dynamic fusion graph for text, audio, and visual modalities with near-state-of-the-art accuracy. Tsai et al.\ proposed IMR (Interpretable Multimodal Routing)~\citep{tsai2020multimodal}, which learns input/output-conditioned routing weights useful for highlighting explanatory modalities; this is the \emph{primary} interpretable baseline in our experiments. State-of-the-art non-interpretable models include PRM~\citep{lv2021progressive}, CBT~\citep{fu2024lmr,fu2021lmrcbt}, and MulT~\citep{tsai2019multimodal}. Other influential architectures include TFN~\citep{zadeh2017tfn}, MFN~\citep{zadeh2018mfn}, multi-attention recurrent networks~\citep{zadeh2018multiattention}, contextual cross-modal attention~\citep{ghosal2018contextual}, and graph-based fusion~\citep{hu2021mmgcn,huddar2020multilevel}.

\section{Proposed Methods}
\label{sec:methods}

\subsection{Discriminative Random Forest, Adaptive Boost, and Gradient Boost}

Embedding LDA-node trees within Random Forest (parallel bagging), AdaBoost, and Gradient Boost (sequential boosting) enables richer multimodal interpretation. Lemmond et al.~\citet{lemmond2010extended} showed that discriminative random forests outperform standard ones~\citep{pal2005random} with lower generalization error. No prior work has applied discriminant projectors to AB and Gradient Boost. Standard oblique trees converge slowly owing to non-convex optimization and impurity-based split bias~\citep{murthy1994system,hada2023sparse}.

\subsection{Proposed Classifier Types: Ensembles of Trees}

\begin{figure}[t]
  \centering
  \includegraphics[width=0.9\linewidth]{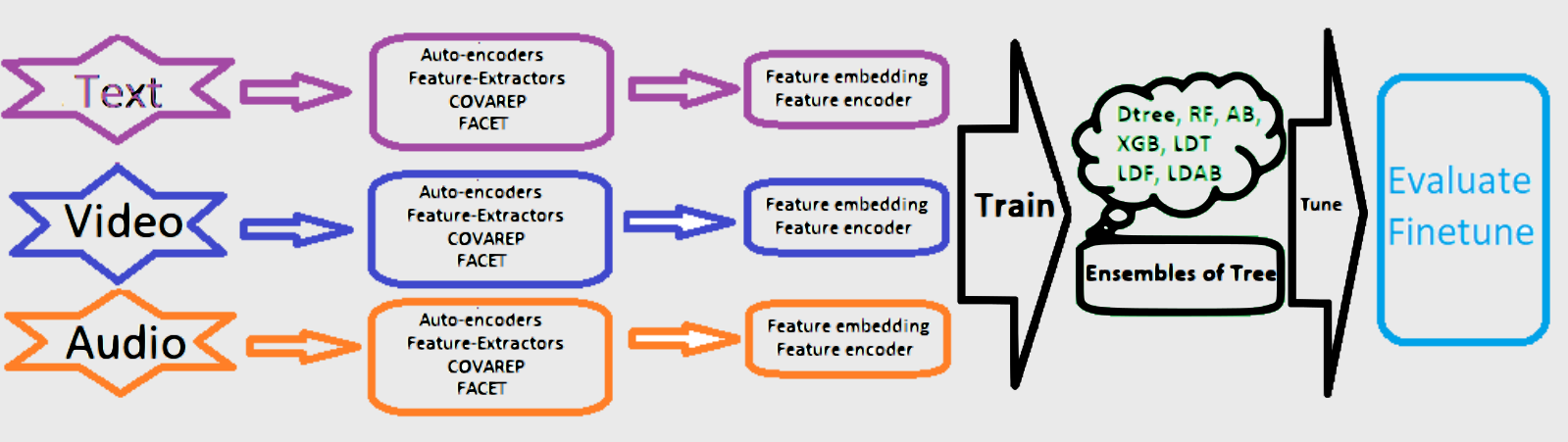}
  \caption{\fontsize{10pt}{11pt}\selectfont{\itshape{Training pipeline for linear-discriminative ensembles of trees.}}}
  \label{fig:training-framework}
\end{figure}

\begin{figure}[t]
  \centering
  \scalebox{0.75}{\includegraphics[width=0.9\linewidth]{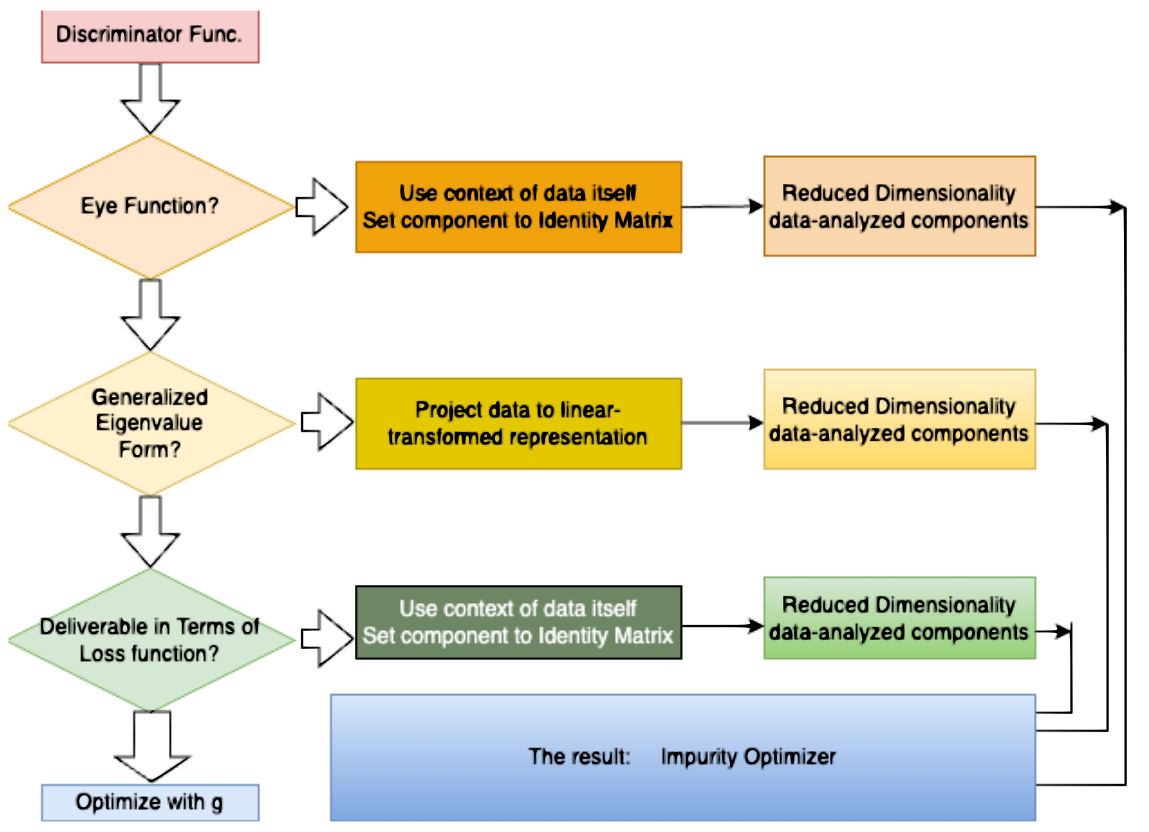}}
  \caption{\fontsize{10pt}{11pt}\selectfont{\itshape{Node discriminator module: decision flowchart for selecting dimensionality-reduction strategy (identity/eye function, generalized eigenvalue projection, or loss-function optimization) at each tree node.}}}
  \label{fig:dim-reduction}
\end{figure}

Standard tree ensembles do not highlight features necessary for positive-label decisions, and conventional feature importance relies on only a subset of nodes~\citep{molnar2020interpretable}. Trees with linear projectors partition the feature space using oblique hyperplanes~\citep{murthy1994system,yildiz2005linear}. Most oblique-tree algorithms minimise impurity without exploiting class-discriminative projectors such as LDA~\citep{hyvarinen2000independent,moattari2025study}. Therefore, we propose a Tree Ensemble framework that supports Linear Discriminant Node projectors and a modified positive-label feature importance. Figures~\ref{fig:training-framework} and~\ref{fig:dim-reduction} show ensemble-tree framework and linear node projector in the trees in the framework respectively.

\subsection{Proposed Framework for Multimodal Tree-Based Classification}
\label{sec:framework}

\begin{figure}[t]
    \centering
    \scalebox{0.75}{\includegraphics[width=0.9\linewidth]{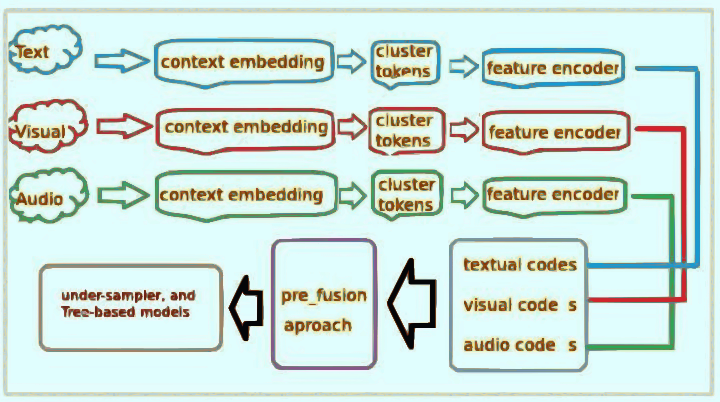}}
    \caption{\fontsize{10pt}{11pt}\selectfont{\itshape{Multimodal classification framework: class balancing, feature coding, and tree-based ensemble classifiers.}}}
    \label{fig:framework-classification}
\end{figure}

The divide-and-conquer inductive bias of Decision Trees yields low estimation bias but becomes computationally intractable for large, high-dimensional datasets. We therefore feed all modalities through a realistic feature encoder to reduce dimensionality~\citep{zhang2010understanding,kim2017bag}. The resulting framework (Figure~\ref{fig:framework-classification}) incorporates class balancing, feature coding, and tree-based ensemble classifiers.

\subsubsection{Context Integration}

For multi-sense clustering, word embeddings are represented using probabilistic multi-sense embeddings~\citep{athiwaratkun2018probabilistic}, enabling context-sensitive token assignment. Eigenvalue-based hierarchical decomposition~\citep{kodinariya2013review} determines the optimal number of sub-clusters per modality. Binary hierarchical clustering encodes each modality's cluster path as a binary-to-decimal integer, so that semantically similar tokens receive numerically adjacent codes, thereby preserving inter-token semantic distance more faithfully than K-means or multi-sense clustering.

\begin{figure}[t]
  \centering
  \includegraphics[width=\linewidth]{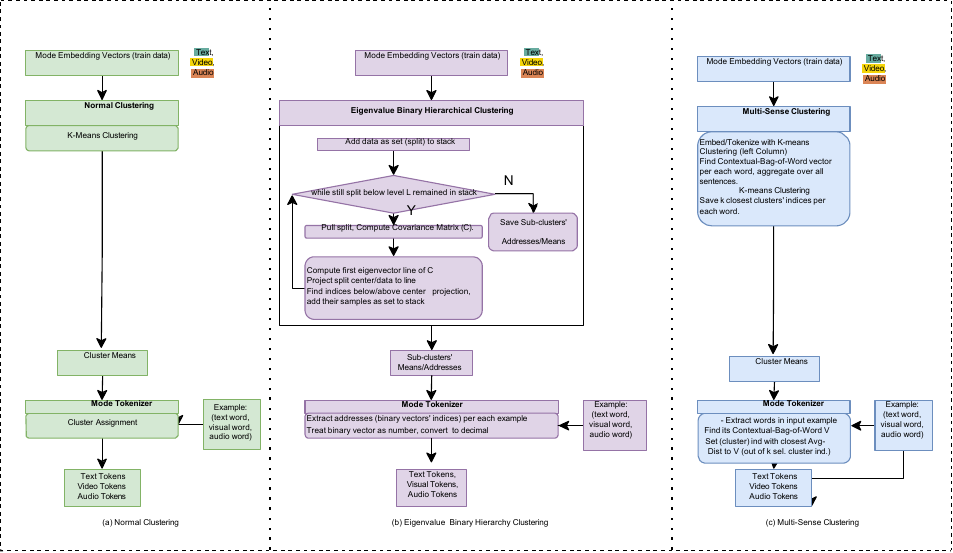}
  \caption{\fontsize{10pt}{11pt}\selectfont{\itshape{Three context-integration approaches: (a) K-means clustering, (b) eigenvalue-based binary hierarchical clustering, and (c) multi-sense clustering. Arrows show token-to-cluster assignment.}}}
  \label{fig:context-integration}
\end{figure}

\subsubsection{Feature Encoder}

After utterances are extracted, the feature encoder produces one of the following representations:
\begin{itemize}
  \item \textbf{Sentence-of-utterance:} Vectorizing statements into sentences for each modality (Table~\ref{tab:fi-math-contexts}).
  \item \textbf{Bag-of-Words (BoW)}~\citep{zhang2010understanding,kim2017bag}: Sparse vectors representing each modality's term frequencies.
  \item \textbf{Bag-of-n-grams}~\citep{chelba2017ngram}: Augments BoW with vectorized upper-triangular bigram matrix, suitable for injecting sequence ordering information in tree ensembles.
\end{itemize}

\subsubsection{Subsampling Methods}

Subsampling addresses class imbalance by removing majority-class instances:
\begin{itemize}
  \item \textbf{Condensed Nearest Neighbour:} Selects a minimal consistent subset.
  \item \textbf{Near-Miss:} Retains majority-class instances farthest from minority-class samples.
  \item \textbf{Edited Nearest Neighbour (Selection):} Removes majority-class instances closest to other classes.
  \item \textbf{Random Subsampling:} Randomly removes majority-class samples.
  \item \textbf{Binary Hierarchical Grouping:} PCA-based hierarchical removal of majority-class samples distant from minority-class neighbours~\citep{kodinariya2013review}.
\end{itemize}

\subsection{Node Projectors}
\label{sec:node-projectors}

Oblique-tree node projectors used in Algorithm~\ref{alg:node-optimizer} include:
\begin{itemize}
  \item Histogram-Based Discriminant Dependency Analysis~\citep{moattari2025study}
  \item Fast Independent Component Analysis~\citep{hyvarinen2000independent}
  \item Linear Discriminant Analysis~\citep{yildiz2005linear}
  \item Principal Component Analysis
\end{itemize}

Although not all projectors exploit class labels, decomposing the feature space into separate positive-class and negative-class subspaces helps identify projections that minimize node impurity.

\subsection{Feature Extractor}

Dedicated feature extraction is employed for audio and visual representations. For affective computing, we use COVAREP~\citep{degottex2014covarep,mcfee2015librosa} for audio and FACET~\citep{imotions2017facet,baltrusaitis2016openface} for video, supplemented by classical descriptors (HOG, PHOG)~\citep{dalal2005hog,zisheng2010phog}.

For the Multimodal Mathematics dataset~\citep{studentperformance}, text was extracted with Tesseract OCR. Each image was processed through quantisation, blurring, dilation, and resizing to produce sparse pixel-intensity values and pixel-location coordinates, each treated as a separate modality.

\begin{figure}[t]
  \centering
  \scalebox{0.5}{\includegraphics[width=\linewidth]{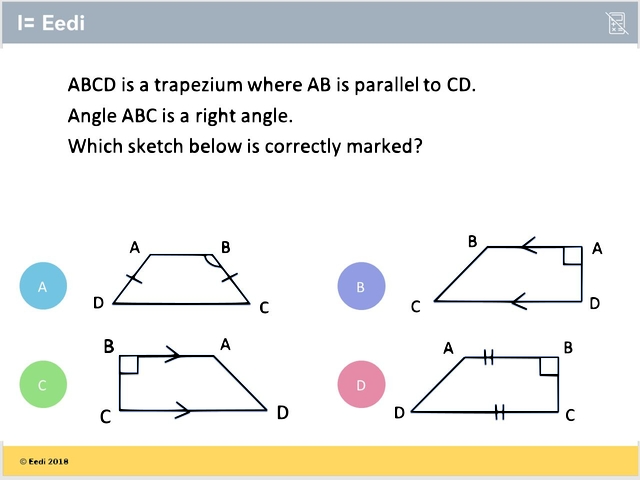}}
  \caption{\fontsize{10pt}{11pt}\selectfont{\itshape{A sample from the Multimodal Mathematics dataset.}}}
  \label{fig:math-sample}
\end{figure}

\subsection{Datasets, Baselines, and Hyper-parameters}
\label{sec:datasets}

We evaluate the proposed framework on three benchmarks.

\textbf{IEMOCAP}~\citep{zadeh2016intensity,poria2017contextdependent} is used for emotion recognition (angry, happy, sad, neutral). \textbf{CMU-MOSI}~\citep{zadeh2018cmu_mosei,zadeh2017tfn} is used for sentiment classification from multimodal data. Both datasets are multi-label and contain textual, visual, and audio modalities per utterance; visual data comprises facial expressions and action units~\citep{kohler2004facial,mehu2015emotioncats,kollias2021affect}. We follow the same approach as~\citep{tsai2019multimodal} for reporting accuracy and F1 scores. For each target label, precision is computed in a one-vs-rest scheme, treating utterances bearing that label as positive and all others as negative. \textbf{Multimodal Maths} is a custom dataset for assessing student performance and identifying strong subject-matter predictors. It contains 20\,000 samples across three modalities: text, geometric shapes (visual), and question/choice images. Label settings LS1--LS4 (shown in Table~\ref{tab:math-results}) are binary thresholds defined as: higher than $m$ (LS1), in the interval $[m - 0.25s,\; m + 0.25s]$ (LS2), higher than $0.5$ (LS3), and higher than $m + 0.25s$ (LS4), where $m$ and $s$ are the sample mean and standard deviation of the number of students who answered each question correctly. Text is extracted by Tesseract OCR; visual and positional data are represented as sparse features extracted with Scikit-Learn.

The primary interpretable baseline is \textbf{IMR} (Interpretable Multimodal Routing)~\citep{tsai2020multimodal}. MulT~\citep{tsai2019multimodal} and LMF-MulT serve as secondary, non-interpretable reference points. All results in Tables~\ref{tab:iemocap-results}, \ref{tab:math-results}, and \ref{tab:cmu-results} are averaged over 10 independent runs. A comprehensive statistical significance test (paired $t$-test with Holm correction versus the IMR baseline) is applied; values marked with $\wedge$ have $p < 0.05$. Entries without $\wedge$ exhibited high cross-run variance; statistical significance relative to IMR could not be established.

\paragraph{Hyper-parameter search} Table~\ref{tab:hyperparams} summarises the search space and final tuned hyper-parameters for each classifier family. A grid-search strategy was used for tree depth and number of estimators; Bayesian optimization
annotator agreement score was applied for LDA regularisation and learning rates.

\begin{table}[t]
  \centering
  \small
  \caption{\fontsize{10pt}{11pt}\selectfont{\itshape{Hyper-parameter search space and final tuned values for each classifier. \textit{Best} denotes the configuration selected by 5-fold cross-validation on the training split.}}}
  \label{tab:hyperparams}
  \renewcommand{\arraystretch}{1.15}
  \scalebox{0.62}{
  \begin{tabular}{l l l l}
    \toprule
    Classifier & Hyper-parameter & Search Space & Best Value \\
    \midrule
    \multirow{3}{*}{Decision Tree (DT)}
      & Max depth      & \{5, 10, 20, None\}    & 20 \\
      & Min samples split & \{2, 5, 10\}         & 5 \\
      & Criterion      & \{gini, entropy\}       & entropy \\
    \midrule
    \multirow{4}{*}{Random Forest (RF)}
      & N estimators   & \{50, 100, 200, 500\}  & 200 \\
      & Max depth      & \{10, 20, None\}        & None \\
      & Max features   & \{sqrt, log2, 0.5\}     & sqrt \\
      & Min samples leaf & \{1, 2, 4\}           & 1 \\
    \midrule
    \multirow{4}{*}{AdaBoost (AB)}
      & N estimators   & \{50, 100, 300\}        & 100 \\
      & Learning rate  & \{0.01, 0.1, 0.5, 1.0\}& 0.1 \\
      & Base estimator depth & \{1, 2, 3\}       & 2 \\
      & Algorithm      & \{SAMME, SAMME.R\}      & SAMME.R \\
    \midrule
    \multirow{5}{*}{XGBoost (XGB)}
      & N estimators   & \{100, 300, 500\}       & 300 \\
      & Max depth      & \{3, 6, 9\}             & 6 \\
      & Learning rate  & \{0.01, 0.05, 0.1\}     & 0.05 \\
      & Subsample      & \{0.6, 0.8, 1.0\}       & 0.8 \\
      & Col sample by tree & \{0.6, 0.8, 1.0\}   & 0.8 \\
    \midrule
    \multirow{3}{*}{LDT / LDF / LDAB}
      & LDA components & \{1, 2, 5, 10\}         & 5 \\
      & LDA shrinkage  & \{None, 0.1, 0.5, auto\}& auto \\
      & Base tree depth & \{5, 10, 20\}           & 10 \\
    \bottomrule
  \end{tabular}}
\end{table}

\subsection{Descriptive Analysis of Auditory Features}

Table~\ref{tab:acoustic-features} (Appendix) describes the 74 acoustic features extracted by COVAREP, an acoustic signal-analysis toolbox. COVAREP processes audio signals and extracts both source-domain and filter-domain features. We design an interpretable approach to identify the most informative action units associated with a speaker's affective state, using cross-modal feature components and the highest-weighted action units.

\subsection{Descriptive Analysis of Visual Features}

We use the FACET tool~\citep{imotions2017facet} for visual feature extraction. Features include facial action units, facial landmarks, head pose, gaze tracking, and HOG descriptors~\citep{dalal2005hog}, sampled at 30\,Hz. In Table~\ref{tab:visual-features} (Appendix), ``*'' denotes emotions that are not exclusively associated with a given action unit. FACET returns a 35-dimensional probability vector per frame; we interpret the top-3 action units. Happy (positive valence) activates rows 3 (Cheek raiser), 15 (Lip corner puller), 28 (Lip pucker), and 30 (Jaw drop). Anger (negative valence) is characterised by rows 1, 2, 4, 17, and 21. Sadness activates rows 16 (Lip corner depressor), 27 (Inner brow raiser), and 32 (Chin raiser), consistent with the low-energy, physically restrained posture characteristic of sadness.

\subsection{Modified Feature Importance}
\label{sec:modified-fi}

\subsubsection{Motivation and Relationship to $F_\beta$}

In medical, educational, and psychological assessment, the key objective is to identify features that indicate \emph{positive} outcomes (e.g., the presence of a target emotion or learning milestone). Standard feature importance assigns high weight to nodes that best separate any pair of classes, regardless of which class is the target, and tends to be dominated by nodes that discriminate the majority (negative) class. This is analogous to the $F_1$ score being agnostic to class asymmetry: just as $F_\beta$ with $\beta > 1$ up-weights recall over precision to emphasise positive-class sensitivity, our modified importance up-weights nodes where positive instances predominate in the right branch. Concretely, the gain for node $j$ is reweighted as:
\begin{equation}
G'_j = G_j \cdot \frac{N_{jrp} - N_{jlp}}{N_{jrp} + N_{jln}} \cdot N_j
\label{eq:modified-gain}
\end{equation}
where the notation is defined as follows (and referenced again in Table~\ref{tab:fi-metric}):
\begin{itemize}
  \item $N_{jrp}$: count of \emph{positive-class} (target) instances routed to the \emph{right} branch at node $j$.
  \item $N_{jln}$: count of \emph{negative-class} instances correctly routed to the \emph{left} (negative) branch at node $j$.
  \item $N_{jlp}$: count of \emph{positive-class} instances routed to the \emph{left} branch at node $j$.
  \item $N_j$: total number of samples reaching node $j$.
  \item $G_j$: the information gain of DT node $j$.
\end{itemize}

The numerator $N_{jrp} - N_{jlp}$ is positive whenever the right branch contains more positive instances than the left, rewarding nodes that successfully concentrate positive cases. The denominator $N_{jrp} + N_{jln}$ normalises by the total number of correctly directed instances (positive to the right, negative to the left), preventing trivially large gains from high-cardinality nodes. The modified feature importance is then:
\begin{equation}
F_i = \frac{\displaystyle\sum_j \mathbb{E}_{f_{j,i}}\, G'_j}{\displaystyle\sum_j G'_j}
\label{eq:modified-fi}
\end{equation}

\subsubsection{F1-mod Metric}

To complement standard F1, we define the modified F1 measure:
\begin{equation}
\text{F1-mod} = \frac{\text{TP}}{\text{TP} + \frac{1}{2}\cdot\frac{\text{FP}}{m} + \frac{1}{2}\cdot\text{FN}}
\label{eq:f1mod}
\end{equation}
where TP, FP, and FN are true positives, false positives, and false negatives respectively, and $m \geq 4$ is the ratio of negative to positive instances in the dataset. This metric is equivalent to the $F_\beta$ measure with $\beta = \sqrt{m/2}$, where the $F_\beta$ formula is $F_\beta = (1+\beta^2)\cdot\text{TP} / [(1+\beta^2)\cdot\text{TP} + \beta^2\cdot\text{FN} + \text{FP}]$. When the negative class is at least four times larger than the positive class, F1-mod penalises false positives less heavily than standard F1, providing a more realistic assessment of model performance under class imbalance. Standard F1 is reported alongside F1-mod in all results tables.

\subsection{Algorithm for General and Local Feature Importance}
\label{sec:alg2}

Algorithm~\ref{alg:fi-multimodal} describes the procedure for computing and visualising both general (global) and local multimodal feature importance using the proposed framework. It relies on the ensemble-tree models trained in the framework of Figure~\ref{fig:training-framework}.

\begin{algorithm}[t]
\caption{Multimodal feature importance: global and local computation}
\label{alg:fi-multimodal}
\begin{algorithmic}[1]
\Require Word, audio, and facial-expression cluster representations;
\Statex \hspace{2.2em}trained ensemble-tree model (from Algorithm~\ref{alg:node-optimizer} and Figure~\ref{fig:training-framework})
\Ensure Unimodal / multimodal feature-importance score plots;
\Statex \hspace{2.2em}averaged classification-probability score plots
\State Extract model parameters from the framework (Figure~\ref{fig:training-framework}).
\State For each concept cluster, extract the corresponding Concept Importance score from the ensemble-tree model using modified FI (Equations~\ref{eq:modified-gain}--\ref{eq:modified-fi}).
\If{local (sample-specific) explanation requested}
  \State Extract the cluster assignment of each unimodal/multimodal input feature.
  \State Retrieve the average and maximum FI score corresponding to that cluster.
  \State Display a per-sample importance visualisation with cluster thickness proportional to FI.
\EndIf
\If{global explanation requested}
  \State Identify the top-ranked unimodal/multimodal/within-context clusters by FI.
  \State \textbf{Text:} visualise by the words closest to the cluster centroid.
  \State \textbf{Audio:} visualise by the top audio features not frequently repeated across the dataset.
  \State \textbf{Visual:} visualise by averaging warped faces within the cluster, using the sub-pipeline: frame resizing $\rightarrow$ pose correction $\rightarrow$ facial landmarking $\rightarrow$ image registration $\rightarrow$ morphing.
\EndIf
\end{algorithmic}
\end{algorithm}

\section{Results}
\label{sec:results}

The classifiers produced by the proposed framework are: Decision Tree (DT), Random Forest (RF), AdaBoost (AB), XGBoost (XGB), Linear Discriminant Tree (LDT), Linear Discriminant Forest (LDF), and Linear Discriminant AdaBoost (LDAB); these acronyms are used in all tables. The \textbf{primary interpretable baseline} throughout is \textbf{IMR} (Interpretable Multimodal Routing~\citep{tsai2020multimodal}); improvements over IMR are the principal measure of interpretable-model progress. MulT and LMF-MulT serve as non-interpretable upper-bound references. Values proposed by this work are shown in \textbf{bold}. All results are averaged over 10 independent experimental runs. Statistical significance is assessed by a paired $t$-test with Holm correction versus IMR; values marked with $\wedge$ satisfy $p < 0.05$. Entries without $\wedge$ exhibited high cross-run variance; significance relative to IMR could not be established.

\subsection{Performance Comparison (RQ1 \& RQ3)}
\label{sec:perf-comparison}

\textbf{Addressing RQ1} (whether tree ensembles can match Transformer accuracy): Tables~\ref{tab:iemocap-results}, \ref{tab:cmu-results}, and~\ref{tab:math-results} show that our proposed ensembles achieve accuracy and F1 competitive with MulT and LMF-MulT, and consistently exceed IMR across the majority of metrics and datasets. For example, LDF achieves F1-mod H of 67.8\% on IEMOCAP, surpassing IMR (50.0\%) by 17.8 percentage points. \textbf{Addressing RQ3} (whether LDA-node projection improves accuracy over standard ensembles): LDT, LDF, and LDAB consistently outperform their projection-free counterparts (DT, RF, AB) on most metrics in Tables~\ref{tab:iemocap-results} and~\ref{tab:math-results}, confirming the benefit of discriminative node projection.

\begin{table}[t]
  \centering
  \small
  \caption{\fontsize{10pt}{11pt}\selectfont{\itshape{IEMOCAP emotion recognition averaged results (10 runs). Models marked * are interpretable by design. Values with $\wedge$ are statistically significant ($p<0.05$, paired $t$-test with Holm correction) versus IMR. Two best results exceeding interpretable models are in \textbf{bold}; two best results overall are \uline{underlined}.}}}
  \label{tab:iemocap-results}
  \scalebox{0.8}{
  \begin{tabular}{lcccccccccc}
    \toprule
    Model & Context Integration & Feature Encoder & Underamp & Acc H & Acc S & F1H & F1S & F1-mod H & F1-mod S \\
    \midrule
    MulT         & --       & GloVe        & --       & 86.5±1.1$^{\wedge}$ & 82.0±4.3 & 84.8±6.6$^{\wedge}$ & 82.0±8.4$^{\wedge}$ & 46.0±6.7 & 66.6±6.5 \\
    LMF-MulT     & --       & GloVe        & --       & 86.4±1.0$^{\wedge}$ & 84.9±4.8$^{\wedge}$ & 84.8±7.1$^{\wedge}$ & 84.6±9.9$^{\wedge}$ & 46.7±5.2 & 69.5±1.4 \\
    IMR*         & --       & GloVe        & --       & 85.6 & 79.4 & 79.0 & 70.3 & 50.0 & 70.3 \\
    \textbf{Ours (RF*)}   & K-means  & BoW(AVT)     & condense & \textbf{86.7}±1.2$^{\wedge}$ & \textbf{81.0}±2.0$^{\wedge}$ & \textbf{82.7}±5.7 & \textbf{79.2}±6.4$^{\wedge}$ & \textbf{60.3}±8.6$^{\wedge}$ & 68.0±3.7 \\
    \textbf{Ours (XGB*)}  & K-means  & BoW+2gram    & --       & \textbf{86.6}±1.2$^{\wedge}$ & \textbf{81.4}±2.8$^{\wedge}$ & \textbf{82.5}±4.5$^{\wedge}$ & \textbf{78.2}±6.3$^{\wedge}$ & \textbf{61.8}±8.8$^{\wedge}$ & 66.8±5.8 \\
    \textbf{Ours (LDT*)}  & K-means  & Arc+2gram    & condense & \uline{\textbf{86.9}}±1.4$^{\wedge}$ & \uline{\textbf{82.4}}±3.8$^{\wedge}$ & \textbf{83.0}±4.5$^{\wedge}$ & \textbf{80.1}±7.7$^{\wedge}$ & \textbf{63.2}±9.6$^{\wedge}$ & 50.8±12.0 \\
    \textbf{Ours (LDF*)}  & K-means  & BoW(AVT)     & Random   & \textbf{86.3}±0.8$^{\wedge}$ & \textbf{81.4}±3.1 & \uline{\textbf{83.5}}±5.0$^{\wedge}$ & \uline{\textbf{80.8}}±7.7$^{\wedge}$ & \uline{\textbf{67.8}}±9.4$^{\wedge}$ & \uline{\textbf{69.5}}±1.3 \\
    \textbf{Ours (LDAB*)} & K-means  & BoW(AVT)     & --       & \textbf{86.0}±0.6 & \textbf{79.8}±0.5$^{\wedge}$ & \textbf{83.5}±6.2$^{\wedge}$ & \textbf{77.2}±5.9$^{\wedge}$ & \textbf{61.0}±8.9$^{\wedge}$ & 62.4±7.5 \\
    \bottomrule
  \end{tabular}}
\end{table}

Among the standard (non-LDA) ensembles, XGBoost achieves the best scores on both conventional and modified metrics. Gradient-based residual tracking improves performance over AdaBoost and Random Forest. The best overall results belong to LDF with a bag-of-bigrams encoder; most metrics exceed all baselines, including non-interpretable models. Classifiers with high standard-F1 scores tend to overfit the training data, yielding lower F1-mod on the test set. LDF achieves the highest F1-mod across all proposed and baseline models. LDAB generalises with high accuracy without requiring under-sampling. LDA-based ensembles (LDT, LDF, LDAB) achieve accuracy comparable to---and in several metrics exceeding---MulT on IEMOCAP, while offering substantially greater interpretability, making them more suitable for fault diagnosis.

\begin{table}[t]
  \centering
  \small
  \caption{\fontsize{10pt}{11pt}\selectfont{\itshape{Multimodal Mathematics dataset: F-measure averaged over 10 runs.}}}
  \label{tab:math-results}
  \begin{tabular}{lccccc}
    \toprule
          & LS1              & LS2              & LS3              & LS4              & \#contexts \\
    \midrule
    LDT  & \textbf{51.7} & 75.8             & 72.6 & 64.8 & 100 \\
    RF   & 49.2 & 79.2 & 76.2 & \textbf{68.2} & 100 \\
    AB   & 44.2 & 81.4 & 76.0             & 65.0 & 100 \\
    XGB  & \textbf{52.3} & 80.0 & 73.6 & 61.7 & 100 \\
    LDAB & 48.3             & \textbf{82.9} & 71.1 & 61.0 & 100 \\
    DT   & 45.4 & 80.7 & \textbf{76.7} & \textbf{70.8} & 200 \\
    LDF  & 48.3 & \uline{\textbf{84.2}} & \textbf{80.8} & 48.3 & 400 \\
    \bottomrule
  \end{tabular}
\end{table}

On the Multimodal Mathematics dataset, XGB and AB achieve the lowest classification scores, while LDT and LDF achieve the highest. Interpretability scores exhibit the inverse pattern: XGB and AB score higher on interpretability, while LDT and LDF score lower. The best classification performance is achieved by LDF (84.2\%), which approaches the level of Multimodal Transformers.

\subsection{CMU-MOSI Results}

\begin{table}[h]
  \centering
  \small
  \caption{\fontsize{10pt}{11pt}\selectfont{\itshape{Binary classification results on CMU-MOSI, averaged over 10 runs. Items with * are interpretable. Values with $\wedge$ are statistically significant versus IMR ($p<0.05$). Entries without $\wedge$ had high cross-run standard deviations (see standard deviation columns) and are therefore not statistically significant relative to IMR.}}}
  \label{tab:cmu-results}
  \begin{tabular}{lccccc}
    \toprule
          & B.A. & F1 & F1-mod & Acc. & Coding \\
    \midrule
    MulT                            & 76.1$\pm$9.8$^{\wedge}$ & 76.4$\pm$9.1$^{\wedge}$ & 62.9$\pm$4.1 & 66.2$\pm$6.9 & GloVe \\
    LMF-MulT                        & 73.8$\pm$8.0$^{\wedge}$ & 73.7$\pm$5.3$^{\wedge}$ & 59.0$\pm$9.9 & 56.2$\pm$9.4 & GloVe \\
    IMR*                            & 67.7 & 68.2 & 65.5 & 74.3 & GloVe \\
    \textbf{Ours (XGBoost)*}        & \uline{72.3$\pm$5.2$^{\wedge}$} & \uline{72.1$\pm$4.8$^{\wedge}$} & \uline{\textbf{71.3$\pm$10.2}}  & \uline{72.1$\pm$3.2} & BoW+2Gram \\
    \textbf{Ours (RF* / BoW+2Gram)} & 69.2$\pm$2.0$^{\wedge}$ & 68.9$\pm$0.7$^{\wedge}$ & \textbf{70.9$\pm$6.3}$^{\wedge}$ & 68.9$\pm$8.5 & BoW+2Gram \\
    \textbf{Ours (LDAB)*}           & 57.3$\pm$16.4 & 60.3$\pm$12.8 & 39.2$\pm$23.7 & 60.3$\pm$14.7 & BoW+2Gram \\
    \textbf{Ours (LDT)*}            & 47.6$\pm$15.8 & 49.4$\pm$15.6 & 61.7$\pm$6.3 & 49.4$\pm$12.6 & BoW+2Gram \\
    \textbf{Ours (LDF)*}            & 62.2$\pm$8.4 & 61.9$\pm$9.4 & 66.0$\pm$0.8 & 61.9$\pm$14.0 & No AV 2Gram \\
    \textbf{Ours (RF* / BoW)}       & 70.2$\pm$2.4$^{\wedge}$ & 69.9$\pm$2.2$^{\wedge}$ & \textbf{70.4$\pm$5.4}$^{\wedge}$ & 69.9$\pm$7.1 & BoW \\
    \bottomrule
  \end{tabular}
\end{table}

Table~\ref{tab:cmu-results} shows CMU-MOSI sentiment results. Values exceeding the IMR baseline are shown in bold. The proposed XGBoost variant achieves the best F1-mod (71.3\%), surpassing IMR (65.5\%) by 5.8 percentage points. Several variants---notably LDAB, LDT, and LDF---do not carry the $\wedge$ marker; this reflects elevated cross-run standard deviations (e.g., LDAB B.A.\ std of $\pm$16.44 and F1-mod std of $\pm$23.76) rather than inferior mean performance, and significance cannot be established under these variance conditions. Overall, the proposed tree-based variants consistently exceed IMR on F1-mod, confirming their suitability for imbalanced sentiment classification.

\subsection{Ablation Study (RQ4)}
\label{sec:ablation}

Ablated blocks include Context Embedder (CE), Feature Encoder (FE), and Node Discriminant Projection (NDP). As shown in Table~\ref{tab:ablation}, removing any block reduces classification performance, confirming that all three components contribute positively.

\begin{table}[t]
  \centering
  \small
  \caption{\fontsize{10pt}{11pt}\selectfont{\itshape{IEMOCAP ablation results for F1/mod-F1 (happy + sad average). Top ablation result per column is shown in \textbf{bold}. Best overall configurations are \uline{underlined}.}}}
  \label{tab:ablation}
  \renewcommand{\arraystretch}{1.1}
  \setlength{\tabcolsep}{3pt}
  \scalebox{0.62}{
  \begin{tabular}{p{3.8cm} c c c c c c c}
    \toprule
    Configuration & DT & RF & AB & XGB & LDT & LDF & LDAB \\
    \midrule
    \textbf{Full (NDP \& CE \& FE)}  & \textbf{76.3}~/~55.2  & \textbf{81.0}~/~62.3  & \textbf{79.0}~/~\textbf{63.8}  & \textbf{80.8}~/~54.8  & \textbf{81.6}~/~57.0  & \uline{82.2~/~\textbf{68.7}}  & 59.8~/~\textbf{54.0} \\
    Without 2-gram                   & 76.1~/~46.4 & 75.9~/~40.5 & 76.9~/~43.8  & 77.2~/~41.2  & 74.9~/~50.9  & 75.05~/~56.6 & 76.5~/~54.3 \\
    Without context integration (CE) & 47.1~/~45.8  & 52.8~/~38.4  & 64.3~/~49.6  & 56.6~/~40.6  & 57.3~/~40.5  & 58.9~/~50.9  & 47.6~/~46.6 \\
    Without functionality coder (FE) & 50.5~/~50.3  & 50.4~/~50.9  & 51.5~/~49.6  & 52.4~/~49.7  & 50.0~/~52.9  & 49.6~/~49.8  & 49.8~/~51.3 \\
    Without downsampler              & 78.2~/~\textbf{63.7}  & 77.4~/~48.5 & 78.9~/~61.1 & 78.9~/~60.6 & 77.9~/~\textbf{72.5}  & 79.5~/~\textbf{71.5} & \textbf{78.9~/~73.6} \\
    Without 2-gram, CE               & 63.1~/~46.6  & 52.1~/~43.1  & 56.3~/~49.7  & 58.7~/~39.9  & 51.0~/~48.2  & 60.5~/~52.6  & 57.2~/~48.8 \\
    Without 2-gram, FE               & 49.5~/~51.1  & 50.9~/~50.2  & 50.0~/~52.6  & 51.0~/~51.2  & 49.9~/~52.1  & 49.9~/~51.0  & 50.8~/~51.5 \\
    Without CE, FE                   & 50.8~/~49.8  & 51.4~/~53.1  & 49.9~/~49.5  & 50.5~/~49.9  & 50.0~/~52.1  & 49.8~/~49.9  & 50.1~/~50.2 \\
    Without 2-gram, FE (alt)         & 51.7~/~52.8  & 49.9~/~50.7  & 51.0~/~51.2  & 50.0~/~50.5  & 49.3~/~50.6  & 53.9~/~50.9  & 50.2~/~51.3 \\
    Without audio-visual             & 73.3~/~56.3 & 70.8~/~62.1 & 75.0~/~48.5 & 74.0~/~49.3 & 71.5~/~50.3 & 74.5~/~48.8 & 70.5~/~58.8 \\
    TF-IDF instead of BoW            & 72.6~/~60.1 & 72.65~/~60.1 & 73.2~/~63.8 & 72.5~/~60.5 & 68.9~/~60.5 & 77.2~/~61.2 & 75.0~/~60.9 \\
    Fisher-ICA instead of LDA (NDP) & 77.1~/~53.6  & 79.3~/~61.4  & 74.9~/~64.8  & 78.5~/~52.2  & 78.0~/~55.8  & 79.3~/~68.7  & 60.6~/~54.7 \\
    PCA instead of LDA (NDP)        & 75.1~/~54.9  & 78.4~/~59.8  & 76.6~/~63.6  & 75.2~/~51.3  & 81.3~/~51.4  & \textbf{82.8}~/~67.5  & 55.4~/~53.0 \\
    Score (top-3 best / rank)        & 77.2~(6th)/~60.0~(6th) & 78.9~(4th)/~61.9~(5th) & 78.1~(5th)/~64.0~(2nd) & 79.4~(3rd)/~58.6~(7th) & 80.3~(2nd)/~63.3~(3rd) & \textbf{81.5~(1st)/~69.6~(1st)} & 76.8~(7th)/~63.0~(4th) \\
    \bottomrule
  \end{tabular}}
\end{table}

\textbf{Addressing RQ4:} The ablation results confirm that context integration (CE) and the feature encoder (FE) both contribute significantly. Removing CE alone drops F1 by up to 29 percentage points; removing FE reduces performance to near chance (49--53\%). This demonstrates that the choice of context-integration scheme---particularly K-means with 2-gram encoding---is essential for accurate and interpretable classification.

Among the full-model variants, LDAB achieves the highest modified F-measure (73.6) in the without-downsampler setting. This is attributable to the complementary information aggregated through boosting combined with the discriminative power of the LDA node projector. LDT and LDF achieve the best modified F-measure values overall. LDF compensates for the over-fitting tendency of individual LDA-node trees by aggregating predictions across the forest, thereby reducing variance and improving robustness to noise, outliers, and out-of-distribution samples.

Among standard
ensembles, boosting generally achieves higher raw accuracy than bagging; however, within the LDA-node family (LDT, LDF, LDAB), boosting variants tend to overfit more readily. Replacing LDA with Fisher-ICA or PCA at the node level reduces performance, confirming that the class-discriminative nature of LDA projection is a key driver of the observed gains.

\subsection{Interpretability in Multimodal Ensemble Classifiers (RQ2)}
\label{sec:interpretability}

For reliable model revision and human-under\-stand\-able explanation, interpretability must improve alongside classification accuracy. Our proposed tree-ensemble framework with modified feature importance is designed to address RQ2: whether modified feature importance leads to measurably better multimodal interpretability. In particular, features that primarily capture negative-class variation can mislead practitioners performing fault diagnosis. We therefore propose modified FI to focus on positive-class indicators.

Few prior studies have addressed interpretation of multimodal data within tree-based ensemble models for binary classification. In such settings, features that are highly selective for rejecting negative instances are frequently conflated with features that are genuinely important for identifying positive ones. The present work provides local and global concept interpretation for multimodal behavioural datasets using the modified feature importance described in Section~\ref{sec:modified-fi}.

\subsubsection{Modified Feature Importance Evaluation Methodology}

In our assessment (Tables~\ref{tab:iemocap-fi} and~\ref{tab:mosi-fi}), human annotators label a dictionary of words from the IEMOCAP database as ``happy'' or ``sad''. We used K-means to group the words into 100 clusters and then labelled each cluster with the majority annotator label. Each cluster is used as an index into the proposed classification framework, and the feature importance for the happy and sad binary classifiers is extracted. Features are assigned a label of $+1$ (happy descriptor) or $-1$ (sad descriptor), and the percentage of accurate label assignments against human-annotated targets is measured. The average and best scores per feature importance variant are shown in Table~\ref{tab:fi-metric}.

\subsubsection{Interpretability Scores Demonstrate Superiority of Modified Feature Importance}

Table~\ref{tab:fi-metric} presents 13 feature importance variants evaluated against human annotation. The proposed Modified FI variant 9 achieves an average agreement score of 45\% and a peak score of 50\%, substantially exceeding Default FI (25\% average, 14\% peak). The results in Tables~\ref{tab:iemocap-fi} and~\ref{tab:mosi-fi} provide further sample-level evidence: modified FI achieves a total score of 24.87/36 on IEMOCAP vs.\ 15.55/36 for default FI---a gain of 60\%---and 16.82/24 vs.\ 11.46/24 on CMU-MOSI---a gain of 47\%. These quantitative improvements demonstrate that the proposed modified feature importance is better suited to multimodal settings than conventional feature importance, as it selectively up-weights nodes where positive-class instances predominate, attenuating the influence of nodes that primarily discriminate the majority negative class. This distinction is particularly important in imbalanced affect datasets where, for example, ``happy'' utterances constitute fewer than 25\% of samples.

Specifically, as shown in Tables~\ref{tab:iemocap-fi} and~\ref{tab:mosi-fi}, the modified FI consistently surfaces emotionally salient words (e.g., \textit{awesome, amazing, wedding, ashamed}) in the top-ranked text features, while default FI frequently selects stop-words or emotionally neutral terms. This pattern holds across audio and visual modalities as well. The single Decision Tree (DT) and AB had the weakest interpretation scores for class~1 (happy). LDT, LDAB, RF, and AB showed sadness interpretation scores closer to the neutral class~3 than to the clearly negative class~2. Overall, XGB, LDF, and RF provide the best explanation of class~1 and class~2 tokens in IEMOCAP. Among all classifiers, LDF is the only model that consistently outperforms the others in interpretation, owing to its subspace-inclusive modified feature importance, which leverages the LDA projection at every node.

\begin{table}[t]
  \centering
  \small
  \caption{\fontsize{10pt}{11pt}\selectfont{\itshape{Designed FI metric variants and their evaluation scores. All formulas assume right-branch nodes carry positive-class instances. The notation $N_{jrp}$, $N_{jln}$, $N_{jlp}$, $N_{j}^{p}$, and $N_j$ is defined in Equation~\ref{eq:modified-gain} and the surrounding text in Section~\ref{sec:modified-fi}. Best average score is \uline{underlined}.}}}
  \label{tab:fi-metric}
  \setlength{\tabcolsep}{4pt}
  \scalebox{0.7}{
  \begin{tabular}{l p{5.0cm} c c p{6.2cm}}
    \toprule
    Variant & $G'_j$ formula & Avg.\ score & Best score & Motivation \\
    \midrule

    Default FI
      & $G_j \cdot N_j$
      & 25 & 14 & Standard average split gain \\

    Modified FI~1
      & $G_j \cdot N_{jrp}$
      & 23 & 65 & Weight nodes by the raw count of positive instances in the right branch \\

    Modified FI~2
      & $G_j \cdot \dfrac{N_{jrp}}{N_{jlp}} \cdot N_j$
      & 21 & 58 & Favour nodes with many positive instances in the right branch and few in the left \\

    Modified FI~3
      & $G_j \cdot \dfrac{N_{jrp}}{N_{j}^{p}} \cdot N_j$
      & 21 & 65 & Favour nodes where the right branch captures a large fraction of all positive instances \\

    Modified FI~4
      & $G_j \cdot \dfrac{N_{jrp}}{N_{jlp}}$
      & 27 & 67 & Relative positive-to-negative ratio in right vs.\ left branch, without sample-count scaling \\

    Modified FI~6
      & $G_j \cdot \dfrac{N_{jrp}}{N_{jrp} + N_{jln} + N_{jlp}} \cdot N_j$
      & 23 & 14 & Fraction of correctly directed instances among all directed instances \\

    Modified FI~7
      & $G_j \cdot (N_{jrp} - N_{jln})$
      & 14 & 29 & Net positive advantage: positive instances directed right minus negative instances directed left \\

    Modified FI~8
      & $G_j \cdot (N_{jrp} - N_{jlp})$
      & 16 & 21 & Surplus of positive instances in the right branch over the left branch \\

    Modified FI~9
      & $G_j \cdot (N_{jrp} - N_{jlp}) \cdot N_j$
      & \uline{45} & 50 & Surplus of positive instances, scaled by node size \\

    Modified FI~10
      & $G_j \cdot \dfrac{N_{jrp} - N_{jlp}}{N_{jrp} + N_{jln}} \cdot N_j$
      & 26 & 51 & Normalised surplus, scaled by node size \\

    Modified FI~11
      & $G_j \cdot (N_{jrp} - N_{jlp}) \cdot N_{jrp} \cdot N_j$
      & 36 & 36 & Surplus weighted by both positive-instance count and node size \\

    Modified FI~12
      & $G_j \cdot (N_{jrp} - N_{jlp}) \cdot \dfrac{N_{jrp}}{N_{jrp}+N_{jln}} \cdot N_j$
      & 40 & 35 & Surplus weighted by normalised positive count and node size \\

    Modified FI~13
      & $G_j \cdot \dfrac{N_{jrp} - N_{jlp}}{1 + N_{jln}} \cdot N_j$
      & 35 & 28 & Surplus penalised by negative-instance count in the left branch \\

    \bottomrule
  \end{tabular}}
\end{table}

Adding linear projector inductive bias to the tree-based ensemble framework improves interpretation power. Adding an ensemble layer on top of a single Decision Tree improves both accuracy and interpretability.

\subsection{Sample-Specific Interpretation with Multimodal Concept Importance (RQ2, Local)}
\label{sec:local-interp}

To mine modality-specific concepts, K-means clustering ($K=1600$) is applied, and each modality is encoded as a Bag-of-Words representation. A unified multimodal embedding is formed by concatenating word embedding vectors with the audio and visual cluster centroids. The three largest features are used for visualisation. Audio and visual descriptions are given in Tables~\ref{tab:acoustic-features} and~\ref{tab:visual-features} (Appendix). Multimodal descriptions are constructed by concatenating text, audio signal characteristics, and visual action units. Clustering ($K=200$) is then applied to the embedding vectors.

\begin{figure}[t]
  \centering
  \includegraphics[width=\linewidth]{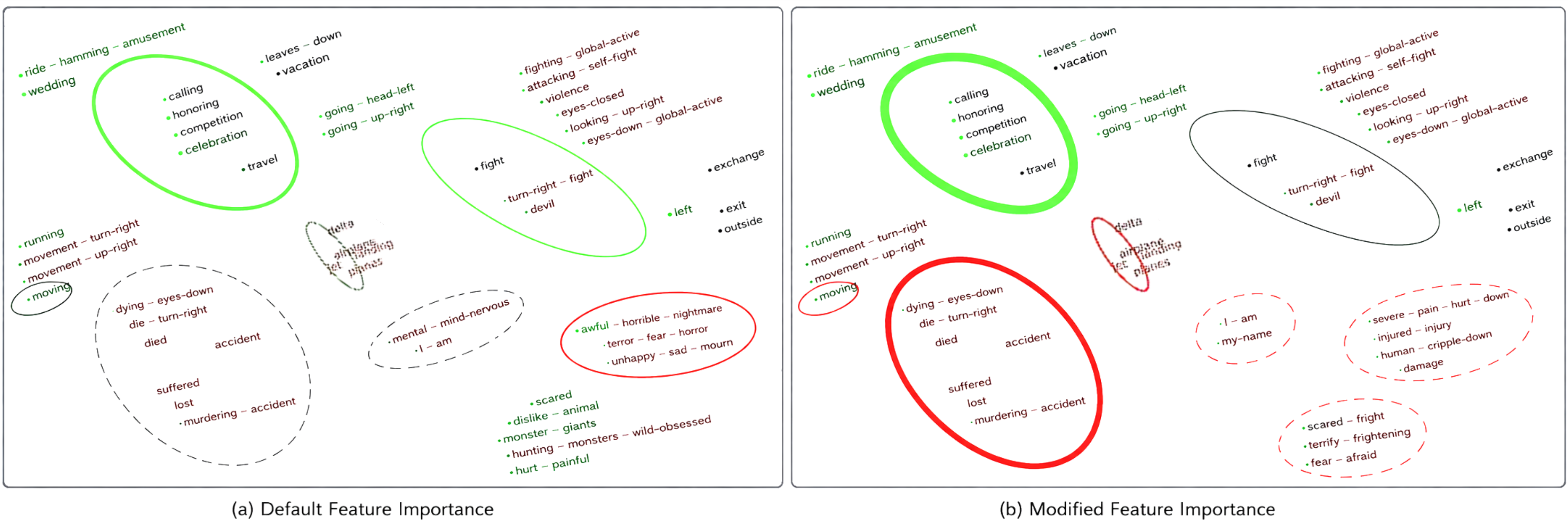}
  \caption{\fontsize{10pt}{11pt}\selectfont{\itshape{Sample-specific interpretation with multimodal concept importance (modified vs.\ default FI). Conventional and proposed local feature importance in Random Forest for multimodal emotions (happy, sad). Each sloping line represents a reduced form of one sentence. Higher-importance clusters have thicker, more intense boundaries; red/green colour indicates higher/lower FI average in the sad vs.\ happy model.}}}
  \label{fig:local-fi}
\end{figure}

\begin{figure}[t]
  \centering
  \includegraphics[width=\linewidth]{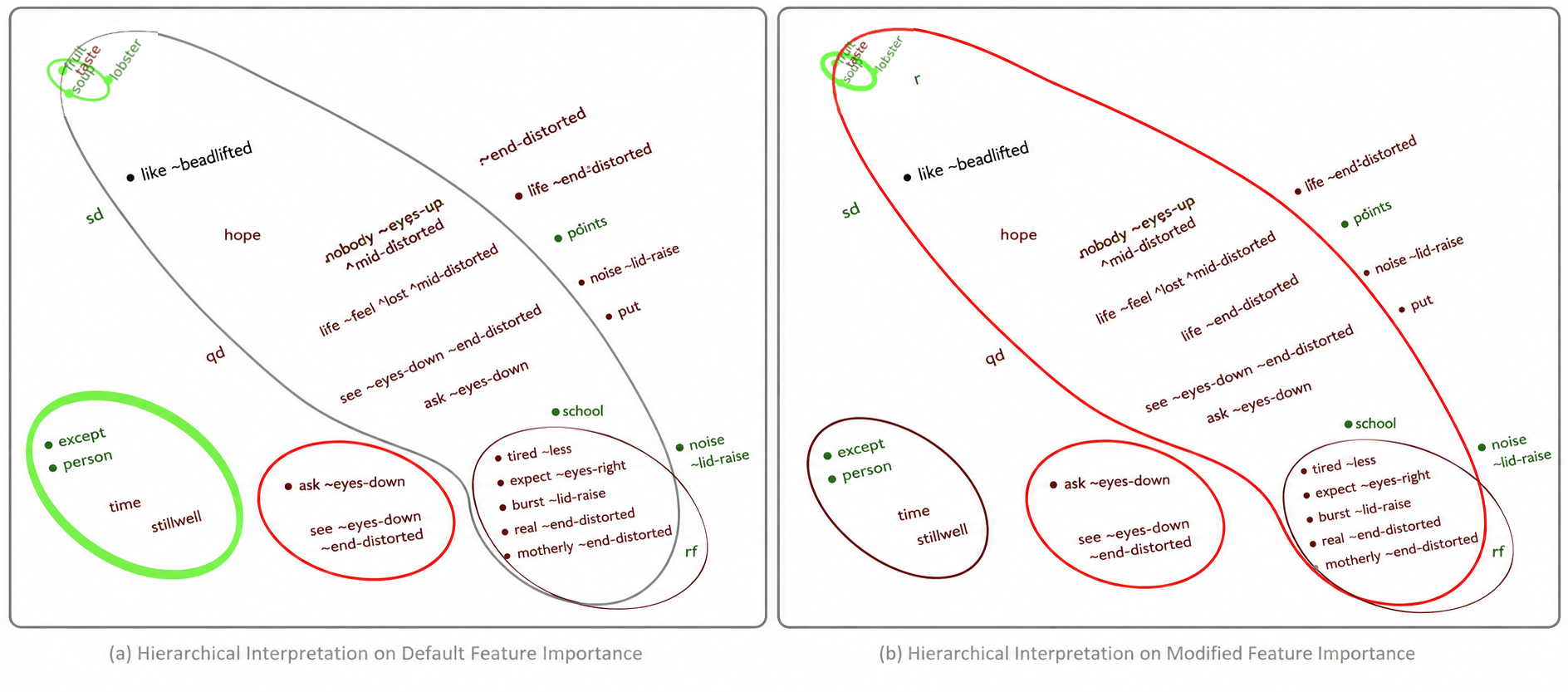}
  \caption{\fontsize{10pt}{11pt}\selectfont{\itshape{Comparison between default and modified feature importance in hierarchical context embedding. Modified FI reduces the importance of emotionally neutral instances (\textit{time}, \textit{person}, \textit{except}, \textit{still}) and produces more realistic polarity colourization of happy and sad groups.}}}
  \label{fig:hierarchical-fi}
\end{figure}

Figures~\ref{fig:local-fi} and~\ref{fig:hierarchical-fi} show sample-specific interpretation results. Two ensemble-tree models are trained---one for the happy binary classifier and one for the sad binary classifier. Higher-importance clusters are displayed with thicker boundaries and more intense colour. Clusters shown in red (green) have higher (lower) average FI in the sad model than in the happy model. Each sloping line represents one sampled instance from the dataset; green text indicates a ground-truth label of ``happy'' and red text indicates ``sad''.

Figure~\ref{fig:local-fi} shows that modified feature importance better delineates the ``happy'' (upper) and ``sad'' (lower) groups, producing greener and thicker boundaries for the happy group. Groups with fewer sentimental instances received thinner and less colourful boundaries under modified FI. Modified FI more successfully identifies the communicative intent behind sentences. For example, the words \textit{delta}, \textit{airplane}, and \textit{jet} appear in red under modified FI, correctly indicating their use in sentences conveying negative emotion; default FI incorrectly labels them as positive. Both methods correctly identify instances with conventionally positive connotations even when used in negative contexts: for example, in the multimodal case ``magically \textasciitilde{}eyes\_down \^{}glottal\_excite'', although \textit{eyes\_down} typically co-occurs with sadness, high glottal excitation frequency leads the interpreter to assign a positive label. The features \textit{harmonics\_variant} and \textit{glottal\_excite} co-occurred with text modality frequently for positive-class samples.

Figure~\ref{fig:hierarchical-fi} shows the same comparison for the hierarchical context-embedding case. Modified FI again reduces the weight assigned to emotionally neutral instances and produces more realistic group colourisation. The proposed hierarchical interpretation can identify subgroups with different majority labels: for example, the words \textit{lobster}, \textit{fruit}, and \textit{soup} appeared in happy discussions, but because their multimodal embeddings are similar to those of most sad instances, K-means interpretation fails to assign the correct polarity. For this case, the modified feature importance identified these words as positive (`happy') with greater certainty (higher green boundary thickness) than default feature importance. Thanks to the hierarchical version of feature importance, these words are identified correctly as happy words even though they are in a larger cluster with `sadness' as the higher-importance label. Had non-hierarchical clustering been used, an incorrect label assignment would result. Both interpreters exploit additional modalities to improve emotion identification: for instance, \textit{eyes\_down} (downward gaze) and \textit{end\_distorted} (increased distortion at the end of an utterance) reliably co-occur with negative affective contexts.

\subsection{General Interpretation in Multimodal Concept Importance (RQ2, Global)}
\label{sec:global-interp}

We define an evaluation measure for comparing interpretation power: the averaged proportion of words conveying the same semantic concept, as judged by human annotators (Table~\ref{tab:fi-math-contexts}). A higher score indicates cleaner semantic grouping. In our assessment (Tables~\ref{tab:iemocap-fi} and~\ref{tab:mosi-fi}), human annotators label a dictionary of IEMOCAP words as ``happy'' or ``sad''. Words are grouped into 100 clusters by K-means and each cluster is labelled with the majority annotator response. These clusters are used as indices into the proposed framework, and the feature importance for the happy and sad binary classifiers is extracted. Each feature receives a label of $+1$ (happy descriptor) or $-1$ (sad descriptor); the percentage of accurate label assignments is measured and averaged across trials.

The results in Tables~\ref{tab:iemocap-fi} and~\ref{tab:mosi-fi} show that adding linear projector inductive bias to tree-based ensembles improves interpretation power. Adding ensemble layers to a single Decision Tree also improves both accuracy and interpretability. We use two multimodal benchmark datasets---IEMOCAP and CMU-MOSI, each with three modalities (audio, visual, text)---to assess and visualise the explanations of the proposed methods.

Tables~\ref{tab:iemocap-fi} and~\ref{tab:mosi-fi} confirm that modified feature importance achieves substantially higher human-annotator agreement scores than conventional feature importance. In almost all cases, modified FI assigns higher importance to emotionally salient features. Single Decision Tree (DT) and AdaBoost (AB) yielded the weakest interpretation scores for the happy class (class~1). LDT, LDAB, RF, and AB showed sadness interpretation scores closer to the emotionally neutral class~3 than to the clearly negative class~2. Overall, the best classifiers for explaining class~1 and class~2 tokens in IEMOCAP are XGB, LDF, and RF. Among all classifiers, LDF is the only model that consistently outperforms the others in interpretation, owing to its subspace-inclusive modified feature importance that leverages the LDA node projection across all tree nodes.

The most accurate classifier among the proposed ensemble tree models typically achieves the lowest interpretability score (Table~\ref{tab:fi-math-contexts}), consistent with a well-known accuracy--interpretability trade-off. Ensemble trees with LDA node projection do not improve total interpretability score on the mathematics dataset but produce more domain-specific, mathematically relevant contexts compared with other ensemble models. Concepts are more semantically salient for LDF and LDT, where the projective inductive bias distributes importance weights more uniformly across all features.

\begin{table}[t]
  \centering
  \small
  \caption{\fontsize{10pt}{11pt}\selectfont{\itshape{FI score for different tree classifiers (IEMOCAP emotion dataset).}}}
  \label{tab:fi-classifiers}
  \renewcommand{\arraystretch}{1.1}
  \scalebox{0.78}{
  \begin{tabular}{l c c p{6.0cm} p{6.0cm}}
    \toprule
    Model & Max FI & Avg FI & ``Positive'' emotions & ``Negative'' emotions \\
    \midrule
    DT   & 60  & 50 & drink, rainbow, colored, laugh, joke, cool, awesome, fun, amusing, funny, laughing, cute, surreal, celebrating, school, college, protection & memorial, grave, monument, funeral, hospital, pain, shaking, sorethroat, bruises, shivering, bother, nonsense, contradict \\
    RF   & 70  & 60 & healthy, academy, profession, educated, supportive, considerate, personable, honest, helpful, happy, lucky, darling, celebrating, excited, honor, wishes, grateful & sudden, worst, avoid, terrible, trouble, weak, accident, poor, emergency, stressful, unfortunate, embarrassed, scared, worried \\
    AB   & 90  & 60 & academy, degree, career, inspire, moments, life, pleasure, love, alive, peace, passion, heart, happiness, memories, joy, soup, fried, living, family, cared, excited, moment, chance, wishes, grateful, congratulations & contradict, ignore, insist, starve, misplace, infuriate, silly, futile, bully, disturb, sting, complicated \\
    XGB  & 90  & 50 & song, music, softest, sweet, pleasure, love, alive, passion, heart, marry, friend, moon, laugh, jokes, amusing, funny, cute, happy, lucky, beach, beautiful, nice, enjoy, wonderful, fascinating, incredible, intimate, cool & report, claim, notice, statement, dares, bully, disturb, sting, hospital, cemetery, funeral, patient, disease, cried, hurt, ashamed \\
    LDT  & 100 & 70 & beautifully, neatly, sweetest, consistently, refreshment, party, holiday, wedding, catering, camping, dinner & expecting, transferred, develop, snake, tire, tempered, burst, scratching, rejection, nonsense, contradict, ignoring, ridiculous, argue, trivial, outrageous, unfair \\
    LDF  & 100 & 40 & glad, woke, care, excited, eventually, fit, quiet, easier, smart, easy, ideal, fast, perfectly, comfortable & pain, shivers, bruising, mistake, bother, unfair, dying, death, grave, monument, funeral, cemetery \\
    LDAB & 100 & 80 & anytime, supportive, fresh, green, garden, grant, satisfactory, acceptance, calm, approved, confirmed, polite, considerate, consistently, granted & dare, sake, biohazard, confine, bully, disturb, debate, issue, error, hit \\
    \bottomrule
  \end{tabular}}
\end{table}

\begin{table}[t]
\centering
\small
\caption{\fontsize{10pt}{11pt}\selectfont{\itshape{IEMOCAP: Sample-level multimodal FI comparison (proposed modified FI vs.\ default FI). Scores reflect human-annotator agreement on positive/negative label assignment.}}}
\label{tab:iemocap-fi}
\scalebox{0.62}{
\begin{tabular}{l l >{\centering\arraybackslash}m{1.6cm} p{3.0cm} c l >{\centering\arraybackslash}m{1.6cm} p{3.0cm} c}
\toprule
\multirow{2}{*}{Emotion} & \multicolumn{4}{c}{Proposed Modified FI} & \multicolumn{4}{c}{Default FI} \\
\cmidrule(lr){2-5}\cmidrule(lr){6-9}
& Text & AU image & Audio & Score & Text & AU image & Audio & Score \\
\midrule
happy & Awesome amazing fantastic & \includegraphics[height=2.43em]{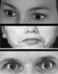} & perceived pitch, harmonic variation, Sect1 resonance & 1+0.66+1 & Pretty cool & \includegraphics[height=2.43em]{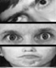} & perceived pitch, harmonic variation, Sect1 resonance & 0.66+0.66+1 \\
happy & Wedding & \includegraphics[height=2.43em]{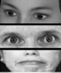} & perceived pitch, Sect4 distortion & 0.33+0.66+0.66 & $<$stopword$>$ & \includegraphics[height=2.43em]{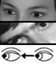} & perceived pitch, Sect3 distortion, Sect4 distortion & 0+0.66+0.33 \\
happy & Beautiful lovely & \includegraphics[height=2.43em]{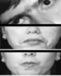} & perceived pitch, Sect1 resonance, Sect4 distortion & 0.66+1+1 & $<$stopword$>$ & \includegraphics[height=2.43em]{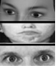} & perceived pitch, Sect4 distortion, Sect1 resonance & 0+0.66+0.33 \\
happy & Sweet taste & \includegraphics[height=2.43em]{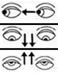} & perceived pitch, glottal variation, Sect1 resonance & 0.66+1+1 & moon & \includegraphics[height=2.43em]{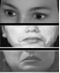} & perceived pitch, Sect1 resonance, peak-glottal-excite & 0.33+0.66+1 \\
sad & Ashamed embarrassed & \includegraphics[height=2.43em]{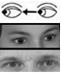} & perceived pitch, Sect1 resonance, Sect3 distortion & 0.66+0.66+0.33 & $<$stopword$>$ & \includegraphics[height=2.43em]{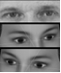} & perceived pitch, Sect4 distortion, Sect4 distortion & 0+0.33+0.66 \\
sad & Meaning & \includegraphics[height=2.43em]{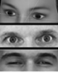} & perceived pitch, glottal variation, Sect1 resonance & 0.33+1+0.33 & Ashamed embarrassed & \includegraphics[height=2.43em]{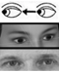} & perceived pitch, Sect1 resonance, Sect3 distortion & 0.66+0.33+0.33 \\
sad & Hard tough & \includegraphics[height=2.43em]{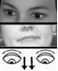} & perceived pitch, glottal variation, Sect1 resonance & 1+1+0.33 & $<$stopword$>$ & \includegraphics[height=2.43em]{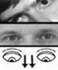} & perceived pitch, Sect1 resonance, Sect4 distortion & 0+0.66+0.33 \\
sad & leave & \includegraphics[height=2.43em]{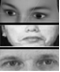} & perceived pitch, Sect1 resonance, Sect2 distortion & 0.33+1+0.33 & $<$stopword$>$ & \includegraphics[height=2.43em]{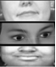} & perceived pitch, harmonic variation, Sect1 resonance & 0+0.33+0.33 \\
angry & Sting intermittently & \includegraphics[height=2.43em]{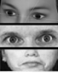} & perceived pitch, harmonic variation, Sect4 distortion & 0.66+0.66+0.33 & $<$stopword$>$ & \includegraphics[height=2.43em]{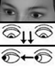} & perceived pitch, Sect1 resonance, Sect4 distortion & 0+1+0.33 \\
angry & shut & \includegraphics[height=2.43em]{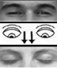} & perceived pitch, harmonic variation, Sect1 resonance & 0.33+1+1 & $<$stopword$>$ & \includegraphics[height=2.43em]{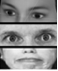} & perceived pitch, Sect4 distortion, Sect3 distortion & 0+0.66+0.66 \\
angry & Beast monster & \includegraphics[height=2.43em]{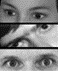} & perceived pitch, harmonic variation, Sect1 resonance & 0.66+0.66+0.66 & business & \includegraphics[height=2.43em]{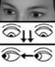} & perceived pitch, Sect1 resonance, harmonic variation & 0+0.66+0.33 \\
angry & Crazy insane & \includegraphics[height=2.43em]{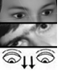} & perceived pitch, Sect4 distortion, Sect3 distortion & 0.66+0.66+0.66 & shut & \includegraphics[height=2.43em]{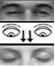} & perceived pitch, harmonic variation, Sect1 resonance & 0.33+1+0.33 \\
\midrule
\multicolumn{2}{l}{Total Score} & \multicolumn{2}{l}{24.87/36} & & \multicolumn{2}{l}{15.55/36} & & \\
\bottomrule
\end{tabular}}
\end{table}

\begin{table}[h]
\centering
\small
\caption{\fontsize{10pt}{11pt}\selectfont{\itshape{CMU-MOSI: Sample-level multimodal FI comparison (proposed modified FI vs.\ default FI).}}}
\label{tab:mosi-fi}
\scalebox{0.62}{
\begin{tabular}{l l >{\centering\arraybackslash}m{1.6cm} p{3.0cm} c l >{\centering\arraybackslash}m{1.6cm} p{3.0cm} c}
\toprule
\multirow{2}{*}{Valence} & \multicolumn{4}{c}{Proposed Modified FI} & \multicolumn{4}{c}{Default FI} \\
\cmidrule(lr){2-5}\cmidrule(lr){6-9}
& Text & Expression & Audio character & Score & Text & Expression & Audio character & Score \\
\midrule
positive & Loved / love & \includegraphics[height=2.88em]{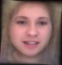} & Sect3 resonance, Sect3 resonance, Sect4 resonance & 0.66+1+1 & $<$stopword$>$ & \includegraphics[height=2.88em]{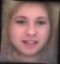} & formant confidence, Sect4 resonance, Sect2 distortion & 0+1+0.66 \\
positive & Perfect / ultimate & \includegraphics[height=2.88em]{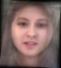} & Sect2 distortion, Sect4 resonance, Sect1 resonance & 0.66+1+0.66 & Impressive / incredible / phenomenal & \includegraphics[height=2.88em]{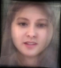} & Sect3 resonance, Sect3 resonance, Sect2 resonance & 1+1+1 \\
positive & Enjoyable / thrilling / exciting & \includegraphics[height=2.88em]{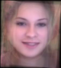} & Sect4 resonance, Sect1 distortion, Sect2 distortion & 1+1+0.33 & Scene / flashback & \includegraphics[height=2.88em]{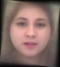} & Sect2 resonance, Sect3 resonance, Sect1 variation & 0+0.5+0.66 \\
positive & Fantastic / wonderful / brilliant & \includegraphics[height=2.88em]{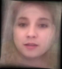} & perceived pitch, Sect4 variation, Sect4 variation & 1+0.5+1 & alright & \includegraphics[height=2.88em]{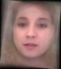} & Sect3 distortion, Sect4 distortion, Sect4 resonance & 0.33+0.5+0.33 \\
negative & terrible & \includegraphics[height=2.88em]{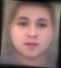} & Sect3 resonance, Sect4 resonance, Sect4 resonance & 0.33+1+0 & funny & \includegraphics[height=2.88em]{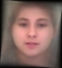} & vocal fold vibration, glottal informative, Sect1 resonance & 0+0.5+0.33 \\
negative & borderline & \includegraphics[height=2.88em]{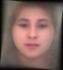} & Sect3 resonance, glottal informative, Sect1 resonance & 0+1+0.33 & disappointed & \includegraphics[height=2.88em]{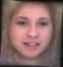} & Sect3 resonance, Sect4 resonance, Sect2 resonance & 0.33+0+0 \\
negative & messy & \includegraphics[height=2.88em]{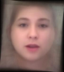} & Sect4 resonance, Sect4 resonance, Sect2 resonance & 0.33+1+0 & $<$stopword$>$ & \includegraphics[height=2.88em]{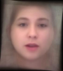} & Sect2 distortion, Sect2 distortion, Sect4 resonance & 0+1+0.66 \\
negative & ridiculous & \includegraphics[height=2.88em]{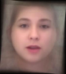} & Sect4 resonance, Sect3 resonance, Sect4 resonance & 0.33+1+0 & action & \includegraphics[height=2.88em]{word/media/image80.png} & peakiest-glottal-excite, formant confidence, Sect4 resonance & 0+1+0.66 \\
\midrule
\multicolumn{2}{l}{Total Score} & \multicolumn{2}{l}{16.82/24} & & \multicolumn{2}{l}{11.46/24} & & \\
\bottomrule
\end{tabular}}
\end{table}

\subsection{General Bimodal Concept Importance: Comparing 1-gram and 2-gram Interpretation}

Rather than encoding groups of concepts from a sentence as a bag-of-words, bigrams encode pairs of adjacent concepts. Table~\ref{tab:bigram-fi} shows the best FI features across all trials. Overall, bigram encoding allows analogy and contextual clarity to be represented more effectively than unigram encoding. LDT, LDF, and LDAB generally produce more emotionally intense word pairs, reflecting the linear subspace of all tree nodes and the aggregate vote across nodes.

\begin{table}[h]
  \centering
  \small
  \caption{\fontsize{10pt}{11pt}\selectfont{\itshape{FI features: comparing 1-gram and 2-gram classifiers.}}}
  \label{tab:bigram-fi}
  \scalebox{0.78}{
  \begin{tabular}{c l p{2.2cm} p{2.2cm} c p{2.5cm}}
    \toprule
    Valence & Classifier & 2-grams (1) & 2-grams (2) & Mode & Compare with 1-gram \\
    \midrule
    + & XGB  & Clasped whispered     & Kiss cheeks              & t     & intimate enjoy cool \\
    + & RF   & Even fantastic        & Wonderful perfect        & t     & honor wishes grateful \\
    + & AB   & Great fantastic brings & Alright okay yeah       & t     & wishes grateful congratulations \\
    + & LDT  & Fascinating incredible amazing & Splendid talented & t     & Beautifully neatly sweetest \\
    + & LDF  & Conga dance           & Dancing music            & t     & excited eventually fit \\
    + & LDAB & Beautiful lovely      & Handsome romantic        & t     & considerate consistently granted \\
    - & XGB  & Ashamed tired bored   & Worried confused angry   & t     & bully hurt ashamed \\
    - & LDT  & Moments mystery wonder & Afraid mystery shocked  & t     & trivial outrageous unfair \\
    - & RF   & Cancer disease        & Mortality risk           & t/v/a & embarrassed scared worried \\
    - & RF   & Trouble lost          & Missing stuck            & t     & embarrassed scared worried \\
    - & AB   & Complicated stressful juggling & Ambiguous difficult frustrating & t & disturb sting complicated \\
    - & LDAB & abrasive              & Conveyor rotary          & t     & issue error hit \\
    \bottomrule
  \end{tabular}}
\end{table}

\begin{table}[t]
  \centering
  \small
  \caption{\fontsize{10pt}{11pt}\selectfont{\itshape{Interpretation score of the Multimodal Mathematics dataset.}}}
  \label{tab:fi-math-contexts}
  \renewcommand{\arraystretch}{1.1}
  \scalebox{0.5}{
  \begin{tabular}{l p{2.2cm} p{2.2cm} p{2.2cm} p{2.2cm} p{2.2cm} p{2.2cm} p{2.2cm} c c}
    \toprule
    Mdl & Context 1 & Context 2 & Context 3 & Context 4 & Context 5 & Context 6 & Context 7 & Score & Rank \\
    \midrule
    LDF & Pyramid out angle point gap diagram (5/6) &
    Sequence parallelines segment (3/3) &
    Odd obtuse chord (1/3) &
    Intersect divide divided (3/3) &
    parallelogram, equilateral, quadrilaterals, rhombus, quadrilateral, parallelograms, decagon, pentagons (8/8) &
    multiple, pairs, identical, units, row, single, sets, fives, pair (6/9) &
    seal, paint, cover, flat, gym, tee, seat, straight, machine, starts, step, eye, board, bucket, balls, seats, club, shoe, fits, gap, slide, track, air (14/22) &
    78.1\% & 7 \\

    LDT & size, fraction, equal, scale, bigger, level, equivalent, cost, gas, total, larger, litre, litres (11/15) &
    Angle, gap, point, ray, diagram (4/5) &
    Sequence paralleline segment tessellate land coordinate (5/5) &
    angle, triangle, rotational, symmetry, parallel, spaced, radii, radius, rotation, rotated (10/10) &
    shape, shapes, form, crystals (4/4) &
    adding, addition, add, fee, additional, item (6/6) &
    crores, lakhs, lakh, eighteen, sixteen (6/6) &
    93.2 & 5 \\

    RF & parallelogram, isosceles, trapezium, rhombus, equilateral, quadrilateral, scalene, decagon (8/8) &
    size, measuring, lengths, shorter, diameter, radius, measurements (7/7) &
    symmetry, associativity, commutativity, distributivity, primes, multiplication, symmetric, integer, finite, infinite, divisible (10/11) &
    diagram, scale, simplified, model, diagrams, estimate, precise, figure, maps, sketch, accurate (7/11) &
    bucket, ale, seat, litre, bottle, lemonade, pours, seats, litres, cola, thermometer, gas, calipers (8/13) &
    earlier, finally, missing, arrive, evening, started, midnight (7/7) &
    box, shoe, pair, nib, set, pairs, sets, boxes, watches, instruments, brackets, machine, item, tapes, pink, instrument, pointer, pieces, fits (14/19) &
    94.0 & 2 \\

    AB & diagonals, diagonal, bisecting, midpoints, sixths, midpoint, intersect, intersects, transversal, equidistant, radii, perpendicular, bisector, fifths (14/14) &
    bucket, ale, seat, litre, bottle, lemonade, pours, seats, litres, cola, thermometer, gas, calipers (14/14) &
    quadrilaterals, triangles, rectangles, vertices, vertex, polygons, polygon, pentagons, parallelograms (9/9) &
    rotated, rotational, rotation, rotate, clockwise, anticlockwise (5/5) &
    size, measuring, lengths, shorter, diameter, radius, measurements (7/7) &
    symmetry, associativity, commutativity, distributivity, primes, multiplication, symmetric, integer, finite, infinite, divisible (11/11) &
    diagram, scale, simplified, model, diagrams, estimate, precise, figure, maps, sketch, accurate (11/11) &
    100 & 1 \\

    XGB & eighteen, quarter, ton, sixes, cows, triple, sixteen, divided, fives (8/8) &
    rectangle, squares, perimeter, shape, dots, octagon, pentagon, triangle, angle, parallel, angles, kite, grid, square, starfish, ascending, shapes, purple, circle, arrow, pyramid, horizontal, vertical, vertically, protractor, ruler, compass, circles, obtuse, cubes, cube, hexagon, dot, crystals, chord (33/36) &
    quadrilaterals, triangles, rectangles, vertices, vertex, polygons, polygon, pentagons, parallelograms (10/10) &
    bucket, ale, seat, litre, bottle, lemonade, pours, seats, litres, cola, thermometer, gas, calipers (12/13) &
    box, shoe, pair, nib, set, pairs, sets, boxes, watches, instruments, brackets, machine, item, tapes, pink, instrument, pointer, pieces, fits (15/18) &
    multiples, sum, fraction, fractions, subtract, multiply, subtracting, percentages, percentage, subtracted, multiplied (11/11) &
    equal, expression, factor, absolute, reflected, represent, values, shares, represented, expressions, reflect, reflection, reflects, divide, expressed, relationship, depths, deepest, represents, sector, positive, relationships, express, equality, transform, measure, draws, negative (28/31) &
    93.9 & 3 \\

    LDAB & symmetry, associativity, commutativity, distributivity, primes, multiplication, symmetric, integer, finite, infinite, divisible (9/11) &
    box, shoe, pair, nib, set, pairs, sets, boxes, watches, instruments, brackets, machine, item, tapes, pink, instrument, pointer, pieces, fits (16/19) &
    parallelogram, isosceles, trapezium, rhombus, equilateral, quadrilateral, scalene, decagon (8/8) &
    arguing, answer, written, write, writes, sentence, statement, statements, quiz, discussing, heed, talking, question, questions, answers, writing, homework, conversation (19/20) &
    rectangle, squares, perimeter, shape, dots, octagon, pentagon, triangle, angle, parallel, angles, kite, grid, square, starfish, ascending, shapes, purple, circle, arrow, pyramid, horizontal, vertical, vertically, protractor, ruler, compass, circles, obtuse, cubes, cube, hexagon, dot, crystals, chord (32/36) &
    multiples, sum, fraction, fractions, subtract, multiply, subtracting, percentages, percentage, subtracted, multiplied (11/11) &
    keeping, sign, friends, arrange, share, money, books, bring, lots, cards (7/10) &
    88.5 & 6 \\

    DT & completed, coordinates, completes, coordinate, finishes, replaces (4/6) &
    quadrilaterals, triangles, rectangles, vertices, vertex, polygons, polygon, pentagons, parallelograms (9/9) &
    temperatures, colder, hotter, degrees (4/4) &
    rectangle, squares, perimeter, shape, dots, octagon, pentagon, triangle, angle, parallel, angles, kite, grid, square, starfish, ascending, shapes, purple, circle, arrow, pyramid, horizontal, vertical, vertically, protractor, ruler, compass, circles, obtuse, cubes, cube, hexagon, dot, crystals, chord (31/34) &
    eighteen, quarter, ton, sixes, cows, triple, sixteen, divided, fives (9/9) &
    pea, tart, chocolate, eats, eat, slices, meal, ate, tuna, lunch (10/10) &
    multiples, sum, fraction, fractions, subtract, multiply, subtracting, percentages, percentage, subtracted, multiplied (11/11) &
    93.9 & 4 \\
    \bottomrule
  \end{tabular}}
\end{table}

The higher the classification accuracy of an ensemble tree model, the lower its interpretability score, consistent with the established accuracy--interpretability trade-off~\citep{molnar2020interpretable}.

\section{Conclusion and Future Work}
\label{sec:conclusion}

\subsection{Summary}

This work proposes tree-based ensemble classifiers---LDT, LDF, and LDAB---for multimodal classification, recognition, and interpretation. The framework integrates context embedding, feature encoding, and LDA-based node projection within a unified pipeline. Concretely:

\begin{itemize}
  \item \textbf{RQ1 (Accuracy vs.\ Transformers):} LDF achieves 86.3\% accuracy and 83.5\% F1 on IEMOCAP happy, and 84.2\% F-measure on Multimodal Mathematics---competitive with, and in some metrics exceeding, MulT~\citep{tsai2019multimodal}.
  \item \textbf{RQ2 (Modified FI):} Modified FI (variant 9) achieves an average interpretability score of 45\% vs.\ 25\% for default FI. At sample level, modified FI scores 24.87/36 on IEMOCAP and 16.82/24 on CMU-MOSI, versus 15.55/36 and 11.46/24 for default FI.
  \item \textbf{RQ3 (LDA projection):} LDA-node variants (LDT, LDF, LDAB) consistently outperform their standard counterparts (DT, RF, AB) on both accuracy and F1-mod, with LDF achieving an F1-mod of 67.8\% on IEMOCAP (happy), surpassing IMR by 17.8 percentage points.
  \item \textbf{RQ4 (Context integration):} Ablation confirms that removing context embedding drops F1 by up to 29 percentage points and that K-means with bigram encoding yields the best balance of accuracy and interpretability.
\end{itemize}

The framework also demonstrates that the proposed modified feature importance method substantially outperforms conventional feature importance in both local and global interpretation tasks, extracting semantically coherent and emotionally salient multimodal concepts across text, audio, and visual modalities.

\subsection{Limitations}

\begin{itemize}
  \item \textbf{Computational cost.} LDA-node projection requires fitting a separate LDA model at each tree node during training, which increases training time relative to standard ensemble methods, particularly for deep trees and large feature spaces. Although inference speed is comparable to standard trees, training may be prohibitive for very large datasets without further optimization
annotator agreement score.
  \item \textbf{Dataset bias.} IEMOCAP and CMU-MOSI are limited in size and demographic diversity. The framework's interpretability benefits have been validated only on English-language, North American speaker populations, and may not generalise to other languages, cultures, or emotional expression norms.
  \item \textbf{Binary classification only.} The modified feature importance is formulated for binary targets. Extension to multi-class settings requires a one-vs.-rest decomposition or a reformulation of the positive-class weighting scheme, which is left for future work.
\end{itemize}

\subsection{Future Work}

Future directions include: improving context representation via adapted autoencoders; adding sparsity constraints to linear node projectors to reduce training cost; exploring dynamic programming for deeper node-interaction optimization
annotator agreement score; and extending the modified FI framework to multi-class and regression settings.

\paragraph{Reproducibility} Code, preprocessed features, trained models, and
evaluation scripts are publicly available at
\url{https://github.com/moatary/reproducible_multimodal_ensemble} to facilitate
replication of all results reported in Tables~\ref{tab:iemocap-results}--\ref{tab:bigram-fi}.

\clearpage
\FloatBarrier
\bibliographystyle{plainnat}
\bibliography{references}

\clearpage
\appendix
\renewcommand{\thesection}{Appendix \Alph{section}}

\section{Acoustic Feature Descriptions}
\label{app:acoustic}

\begin{table}[t]
  \centering
  \small
  \caption{\fontsize{10pt}{11pt}\selectfont{\itshape{The 74 acoustic features extracted by COVAREP, organized by feature group (Pitch/source, Glottal/source, Phase/source, Amplitude/ filter, Phase/filter). Column `R' denotes the feature index used in the main text.}}}
  \label{tab:acoustic-features}
  \renewcommand{\arraystretch}{1.1}
  \scalebox{0.5}{
  \begin{tabular}{l c l l p{2.2cm} p{2.0cm} p{1.6cm} p{3.2cm}}
    \toprule
    Group & R & Feature & Caption Name & Description & Emotion & Sentiment & Extraction Mechanism \\
    \midrule
    Pitch/source    & 1  & F0           & Perceived pitch                 & Perceived fundamental frequency                              & Happy, Angry    & Positive valence                  & Frequency with highest summed residual harmonics (SRH) \\
    Glottal/source  & 2  & VUV          & Voicedness                      & Sound produced by laryngeal vibration                        & Angry, Happy                  & Negative/positive valence         & Percentage of segments exceeding the SRH threshold \\
    Glottal/source  & 3  & NAQ          & Glottal informativeness         & Informativeness of glottal pattern                           & Non-neutral                   & --                                & Normalised amplitude quotient \\
    Glottal/source  & 4  & QOQ          & Vocal fold vibration            & Duration of open phase relative to glottal cycle             & Non-neutral                   & Negative valence                  & Open-phase duration divided by glottal cycle duration \\
    Glottal/source  & 5  & H1H2         & Glottal harmonic variation      & Spectral variation of voice source                           & Angry                         & Negative valence                  & Difference between peak frequencies of the two highest spectral peaks \\
    Glottal/source  & 6  & PSP          & Glottal volume velocity         & Quantification of glottal volume velocity                    & Happy                         & Positive valence                  & Parabolic Spectral Parameter (PSP) \\
    Glottal/source  & 7  & MDQ          & Glottal excitation impulse      & Degree of pressed phonation (tightly adducted glottis)       & Anger, Non-neutral            & Negative valence                  & Wavelet classification of LP residual to measure glottal closure dispersion \\
    Glottal/source  & 8  & peakSlope    & Voice laxness                   & Variation in glottal formant peak values                     & Sadness, Non-neutral          & --                                & Estimating and comparing glottal source peaks per frequency band \\
    Phase/source    & 9  & Rd           & Glottal excitation shape        & Deterministic component of the glottal source                & Non-neutral                   & Positive valence                  & Inverse filtering for parametric glottal source representation \\
    Phase/source    & 10 & Rd\_conf      & Formant confidence              & Confidence in the glottal excitation estimate                 & Non-neutral                   & --                                & Variational confidence measure \\
    Glottal/source  & 11 & creak        & Voice creak                     & Degree of tense vocal-fold excitation (vocal/glottal fry)    & Anger, Sadness                & Positive valence                  & Decision tree classifying segment as creaky/non-creaky \\
    Amplitude/filter & 12 & MCEP\_0:5   & Section 1 resonance             & Resonance in the first spectral quarter                      & Happy, Angry, Sadness         & Positive valence, phone-dep.      & Mel cepstral computation for spectral envelope parameterisation \\
    Amplitude/filter & 13 & MCEP\_6:11  & Section 2 resonance             & Resonance in the second spectral quarter                     & Happy, Angry, Sadness         & Positive valence, phone-dep.      & Mel cepstrum: non-minimum and minimum-phase decomposition \\
    Amplitude/filter & 14 & MCEP\_12:17 & Section 3 resonance             & Resonance in the third spectral quarter                      & Happy, Angry, Sadness         & Positive valence, phone-dep.      & Mel cepstrum: non-minimum and minimum-phase decomposition \\
    Amplitude/filter & 15 & MCEP\_18:24 & Section 4 resonance             & Resonance in the fourth spectral quarter                     & Happy, Angry, Sadness         & Positive valence, phone-dep.      & Mel cepstrum: non-minimum and minimum-phase decomposition \\
    Phase/filter    & 16 & HMPDM\_0:5  & Section 1 phase distortion      & Phonation irregularity in segment 1                          & Surprise, Happy               & Phone-dependent                   & Averaging phase distortion to convey formant envelope \\
    Phase/filter    & 17 & HMPDM\_6:11 & Section 2 phase distortion      & Phonation irregularity in segment 2                          & Surprise, Happy               & Phone-dependent                   & Averaging phase distortion; section = 2nd temporal quarter \\
    Phase/filter    & 18 & HMPDM\_12:17 & Section 3 phase distortion     & Phonation irregularity in segment 3                          & Surprise, Happy               & Phone-dependent                   & Averaging phase distortion; section = 3rd temporal quarter \\
    Phase/filter    & 19 & HMPDM\_18:24 & Section 4 phase distortion     & Phonation irregularity in segment 4                          & Surprise, Happy               & Phone-dependent                   & Averaging phase distortion; section = 4th temporal quarter \\
    Phase/filter    & 20 & HMPDD\_0:5  & Section 1 phase distortion var. & Phonation irregularity variance in segment 1                 & Surprise, Happy               & Phone-dependent                   & Variance of phase distortion in the 1st temporal segment \\
    Phase/filter    & 21 & HMPDD\_6:12 & Section 2 phase distortion var. & Phonation irregularity variance in segment 2                 & Surprise, Happy               & Phone-dependent                   & Variance of phase distortion in the 2nd temporal segment \\
    \bottomrule
  \end{tabular}}
\end{table}

\section{Visual Feature Descriptions}
\label{app:visual}

\begin{table}[t]
  \centering
  \small
  \caption{\fontsize{10pt}{11pt}\selectfont{\itshape{Visual (FACET) features: facial Action Units (AUs), associated facial expressions, emotions, and sentiment valence. `*' denotes emotions that are not exclusively associated with the listed action unit (i.e., the AU also occurs in other emotional contexts). Column `\#AU' gives the FACS AU number.}}}
  \label{tab:visual-features}
  \renewcommand{\arraystretch}{1.3}
  \scalebox{0.5}{
  \begin{tabular}{l c l >{\centering\arraybackslash}m{1.8cm} c l l}
    \toprule
    Grp & \# & Action Unit (AU) & Expression & \#AU & Emotion & Sentiment (Valence) \\
    \midrule
    eye & 1  & Brow lowerer         & \includegraphics[height=1.4em]{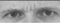}  & 4  & Sadness, Anger, Fear                              & Negative \\
    eye & 2  & Upper lid raiser     & \includegraphics[height=1.4em]{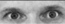}  & 5  & Anger, Surprise, Fear, Happiness*                 & Context dependent \\
    eye & 3  & Cheek raiser         & \includegraphics[height=1.4em]{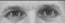}  & 6  & Happiness, Sadness*                               & Positive \\
    eye & 4  & Lid tightener        & \includegraphics[height=1.4em]{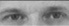}  & 7  & Anger, Fear, Happiness*, Sadness*                 & Negative \\
    eye & 5  & Lid droop            & \includegraphics[height=1.4em]{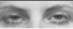} & 41 & Context dependent                                 & Context dependent \\
    eye & 6  & Slit                 & \includegraphics[height=1.4em]{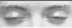} & 42 & Context dependent                                 & Context dependent \\
    eye & 7  & Eyes closed          & \includegraphics[height=1.4em]{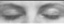} & 43 & Context dependent                                 & Positive \\
    eye & 8  & Squint               & \includegraphics[height=1.4em]{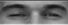} & 44 & Context dependent                                 & Context dependent \\
    eye & 9  & Blink                & \includegraphics[height=1.4em]{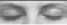} & 45 & Context dependent                                 & Context dependent \\
    eye & 10 & Eyes turn left       & \includegraphics[height=1.4em]{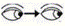} & 61 & Context dependent                                  & Context dependent \\
    eye & 11 & Eyes turn right      & \includegraphics[height=1.4em]{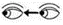} & 62 & Context dependent                                  & Context dependent \\
    eye & 12 & Eyes up              & \includegraphics[height=1.4em]{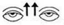} & 63 & Context dependent                                  & Context dependent \\
    eye & 13 & Eyes down            & \includegraphics[height=1.4em]{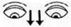} & 64 & Context dependent                                  & Context dependent \\
    lip & 14 & Upper lip raiser     & \includegraphics[height=1.4em]{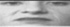} & 10 & Happiness*                                        & Negative \\
    lip & 15 & Lip corner puller    & \includegraphics[height=1.4em]{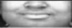} & 12 & Happiness, Contempt                               & Positive \\
    lip & 16 & Lip corner depressor & \includegraphics[height=1.4em]{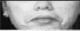} & 15 & Sadness, Disgust                                  & Negative \\
    lip & 17 & Lower lip depressor  & \includegraphics[height=1.4em]{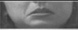} & 16 & Anger                                             & Negative \\
    lip & 18 & Lip funneler         & \includegraphics[height=1.4em]{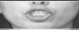} & 22 & Context dependent                                 & Context dependent \\
    lip & 19 & Lip stretcher        & \includegraphics[height=1.4em]{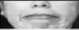} & 20 & Fear                                              & Context dependent \\
    lip & 20 & Lip pucker           & \includegraphics[height=1.4em]{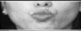} & 18 & Context dependent                                 & Positive \\
    lip & 21 & Lip tightener        & \includegraphics[height=1.4em]{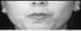} & 23 & Anger                                             & Context dependent \\
    lip & 22 & Lips part            & \includegraphics[height=1.4em]{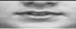} & 25 & Anger*, Happiness*                                & Negative \\
    lip & 23 & Lip suck             & \includegraphics[height=1.4em]{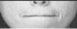} & 28 & Context dependent                                 & Context dependent \\
    head & 24 & Head turn left      & \includegraphics[height=1.4em]{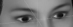} & 51 & Context dependent                                  & Context dependent \\
    head & 25 & Head turn right     & \includegraphics[height=1.4em]{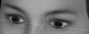} & 52 & Context dependent                                  & Context dependent \\
    head & 26 & Head tilt right     & \includegraphics[height=1.4em]{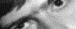} & 56 & Context dependent                                  & Context dependent \\
    other & 27 & Inner brow raiser  & \includegraphics[height=1.4em]{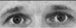} & 1  & Sadness, Surprise, Fear                           & Negative \\
    other & 28 & Outer brow raiser  & \includegraphics[height=1.4em]{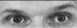} & 2  & Surprise, Fear                                    & Positive \\
    other & 29 & Nose wrinkler       & \includegraphics[height=1.4em]{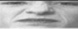} & 9  & Disgust, Anger                                    & Negative \\
    other & 30 & Jaw drop            & \includegraphics[height=1.4em]{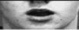} & 26 & Surprise, Fear, Happiness*                        & Positive \\
    other & 31 & Dimpler             & \includegraphics[height=1.4em]{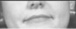} & 14 & Contempt                                          & Negative \\
    other & 32 & Chin raiser         & \includegraphics[height=1.4em]{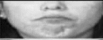} & 17 & Sadness, Disgust, Anger*                          & Negative \\
    \bottomrule
  \end{tabular}}
\end{table}

\end{document}